\documentclass[10pt, journal, compsoc]{IEEEtran}
\usepackage{amsmath,amsfonts}
\usepackage{array}
\usepackage[caption=false,font=footnotesize,labelfont=rm,textfont=rm]{subfig}
\usepackage{textcomp}
\usepackage{stfloats}
\usepackage{url}
\usepackage{verbatim}
\usepackage{graphicx}
\usepackage{booktabs}
\usepackage{subfig}
\usepackage{multirow}
\usepackage[linesnumbered,ruled,vlined]{algorithm2e}
\usepackage{cite}
\usepackage{soul, xcolor}
\usepackage{color, colortbl}
\definecolor{Gray}{rgb}{0.9, 0.9, 0.9}

\usepackage[colorlinks, linkcolor=blue]{hyperref}
\begin{document}

\title{FedIN: Federated Intermediate Layers Learning for Model Heterogeneity}

\author{Yun-Hin Chan, Zhihan Jiang, Jing Deng and Edith C. H. Ngai*

\thanks{Yun-Hin Chan, Zhihan Jiang, Jing Deng and Edith C. H. Ngai (corresponding author) are with the Department of Electrical and Electronic Engineering, The University of Hong Kong, Pokfulam, Hong Kong (e-mail: chanyunhin@connect.hku.hk; zhjiang@connect.hku.hk; u3008395@connect.hku.hk; chngai@eee.hku.hk).}
}

\IEEEtitleabstractindextext{
\begin{abstract}
Federated learning (FL) facilitates edge devices to cooperatively train a global shared model while maintaining the training data locally and privately. However, a common assumption in FL requires the participating edge devices to have similar computation resources and train on an identical global model architecture. 
In this study, we propose an FL method called Federated Intermediate Layers Learning (FedIN), supporting heterogeneous models without relying on any public dataset. Instead, FedIN leverages the inherent knowledge embedded in client model features to facilitate knowledge exchange. 
The training models in FedIN are partitioned into three distinct components: an extractor, intermediate layers, and a classifier. We capture client features by extracting the outputs of the extractor and the inputs of the classifier.
To harness the knowledge from client features, we propose IN training for aligning the intermediate layers based on features obtained from other clients. IN training only needs minimal memory and communication overhead by utilizing a single batch of client features.
Additionally, we formulate and address a convex optimization problem to mitigate the challenge of gradient divergence caused by conflicts between IN training and local training.
The experiment results demonstrate the superior performance of FedIN in heterogeneous model environments compared to state-of-the-art algorithms. Furthermore, our ablation study demonstrates the effectiveness of IN training and the proposed solution for alleviating gradient divergence.
\end{abstract}


}

\maketitle

\section{Introduction}

The massive increase in the usage of Internet-of-Things (IoT) devices induces enormous amounts of data from users \cite{8994077}. Managing these IoT big data efficiently without invading user privacy becomes a significant concern. \textbf{Federated Learning} (FL) \cite{mcmahan2017communication} is proposed as a distributed machine learning paradigm that enables collaborative training on the IoT data while keeping the user data locally. FedAvg \cite{mcmahan2017communication} is the first training paradigm introducing the details of FL. All the clients transmit weights from their local models to the server after a few local training epochs. The server averages these weights to update the global model and sends this model back to clients. 

Although FL has been employed successfully in many applications, such as recognizing human activities \cite{chen2019communication} and learning sentiment \cite{smith2017federated}, many practical challenges of FL still remain to be solved \cite{kairouz2019advances}. 
One of the most crucial and practical challenges is system heterogeneity, which is usually described as different available resources of the client devices in the FL training process \cite{li2020federated}. Many existing FL schemes (e.g., FedAvg) assume that the client devices with distinct resources possess the same architecture as the global shared model for global aggregation. Nevertheless, some clients with lower computation resources may be unable to complete their local training in time, dragging the training speed of the entire communication round. The clients hindering the training process are called stragglers. Some research has proposed asynchronous FL \cite{xie2019asynchronous,chen2020asynchronous,10.1145/3458817.3476211}, maintaining the adaptive local training epochs for clients and clustering clients according to their available resources in order to mitigate the problem of stragglers.
Nonetheless, given that all clients keep the same model architecture, it is still possible that less capable clients do not have sufficient memory to deploy the shared global model. 
In this case, the global model must be adjusted to a smaller size, leading to the resource waste of more capable clients and reducing the performance of FL training.
In fact, it is impractical to guarantee that all the clients have the same amount of resources, particularly in IoT systems. It is common to have heterogeneous devices with different capabilities working together. Thus, supporting heterogeneous models could fully utilize the resources of heterogeneous devices and better address the system heterogeneity problem.

\begin{figure}[!t]

\centering
\includegraphics[width=0.45\textwidth]{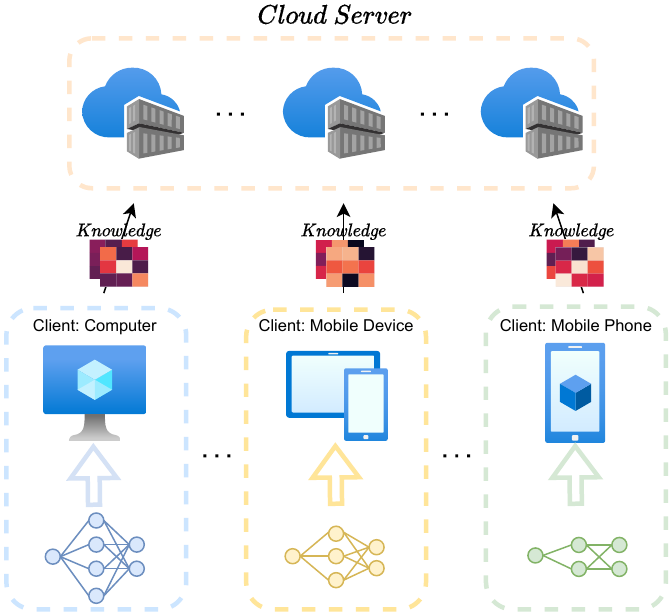}%
\caption{\textbf{An illustration for model heterogeneity.} The clients participate in the federated learning with different available resources, inducing different model architectures.}
\label{fig_sys_hetero}
\end{figure}

A straightforward way to facilitate system heterogeneity is to deploy different model architectures based on the available resources of the clients, as shown in \figurename~\ref{fig_sys_hetero}. However, the server can not aggregate the weights directly like FedAvg under the heterogeneous model architectures. It is essential to investigate alternative ways to incorporate weights and knowledge among the clients. Some recent works propose to solve this problem by performing knowledge distillation \cite{hinton2015distilling} on a public dataset, such as RHFL \cite{fang2022robust} and FedMD \cite{li2019fedmd}. Although they can set up various model architectures in the clients, it is challenging to collect a valid public dataset with a similar distribution to the users' local datasets in practice.
To address this issue, inspired by model parallelism-based split learning \cite{gupta2018distributed, vepakomma2018split}, FedGKT \cite{he2020group} divides the model architectures into two parts, collecting features from the outputs of the former part of models in the clients and training the remaining model in the server based on the collected features.
HeteroFL \cite{diao2021heterofl} is another line of work to support heterogeneous models without relying on a public dataset. It derives local models with different sizes from the largest model (considered to be the server model). However, this approach still requires all client models to share the same architecture design just with different sizes.

To support system heterogeneity, we propose a method called Federated Intermediate Layers Learning (\textbf{FedIN}), training the intermediate layers according to one batch of the collected features from other clients. In FedIN, a local model architecture consists of three components: an extractor, intermediate layers, and a classifier, as depicted in \figurename~\ref{fig_FedIN_process}. Client features are derived from the outputs of the extractor and the inputs to the classifier.
Notably, clients only need to transmit \textbf{one batch} of features to the server, in addition to weight updates.
The intermediate layers are updated through a combination of local training and \textbf{IN} training process, where IN training leverages a single batch of features to extract knowledge from other clients.
However, directly deploying these two training processes can induce a critical problem called gradient divergence \cite{wang2020tackling, zhao2018federated} because the latent information from the local dataset and the features collected from other clients varies. To alleviate the effect of this problem, we formulate and solve a convex optimization problem to obtain the optimal updated gradients.
The experiment results demonstrate that FedIN outperforms the baselines in terms of both accuracy and overheads.

Our contributions are summarized as follows.
\begin{itemize}
    \item We proposed a novel FL method called \textbf{FedIN}, utilizing local training and IN training for intermediate layers, which is a flexible and reliable FL method addressing the system heterogeneity problem. 
    \item To alleviate the effects of the gradient divergence, we formulate a convex optimization problem to derive the optimal updated gradient. The ablation study shows its effectiveness in handling the gradient divergence problem. 
    \item Our experiments reveal that FedIN achieves the best performances in the IID and non-IID data compared with the state-of-the-art algorithms. Moreover, we conduct a thorough analysis to investigate the factors contributing to the improvements attained by FedIN.
\end{itemize}

\section{Related work}

In this section, we first introduce federated learning and then review and classify the works for model heterogeneity into three categories.

\subsection{Federated Learning}

Federated Learning (FL) was proposed by Google in 2017 to organize cooperative model training among edge devices and servers \cite{mcmahan2017communication}. In FL, numerous clients train models jointly while retaining training data locally to maintain privacy protection. Various methods have been proposed and achieved good performance in different scenarios. 
In \cite{xie2019asynchronous}, FedAsyn utilizes coordinators and schedulers to create an asynchronous training process, handling the stragglers in the FL training process. 
FedProx \cite{li2018federated} regularizes and re-parametrizes FedAvg, guaranteeing convergence when learning over non-IID data. 
To share local knowledge among clients with different model architectures, FCCL \cite{huang2022learn} generates a cross-correlation matrix based on the unlabeled public dataset.
In this work, we propose a novel FL framework called FedIN, which supports the training of heterogeneous models at edge devices with heterogeneous resources.

\subsection{Heterogeneous Models}
Our work focuses on supporting heterogeneous models in FL. This subsection classifies recent research contributing to model heterogeneity into three categories.

\subsubsection{Public and auxiliary data}
If a server has a public dataset, clients can exploit the general knowledge from this dataset, constructing a simple and efficient bridge to exchange knowledge among clients.
FedAUX \cite{sattler2021fedaux} utilizes unsupervised pre-training and unlabeled auxiliary data to initialize heterogeneous models.
FedGen \cite{zhu2021data} simulates the prior knowledge from all the clients according to a generator. 
To dig out the latent knowledge from the public dataset, several studies \cite{li2019fedmd,li2021fedh2l,he2020group} are proposed to address the system heterogeneity problem, inspired by the knowledge distillation. Specifically,
knowledge distillation (KD) \cite{hinton2015distilling} was proposed by Hinton et al., training a student model with the knowledge distilled from a teacher model. 
In FedMD \cite{li2019fedmd}, a large public dataset is deployed in a server, while the clients distill and transmit logits from this dataset to learn the knowledge from both logits and local private datasets. 
In FedH2L \cite{li2021fedh2l}, clients extract the logits from a public dataset consisting of small portions of local datasets from other clients.
In RHFL \cite{fang2022robust}, a server calculates the weights of clients by the symmetric cross-entropy loss function, and clients distilled knowledge from the unlabeled dataset. 
FCCL \cite{huang2022learn} computed a cross-correlation matrix also based on the unlabeled public dataset.

\subsubsection{Data-free knowledge distillation}
However, the former methods using KD in FL acquire a public dataset. The server may not collect sufficient data due to data availability and privacy concerns.
In contrast to the aforementioned methods, data-free KD is a novel approach to complete the knowledge distillation process without the training data. The basic ideas of data-free KD are to optimize noise inputs to minimize the distance to prior knowledge \cite{nayak2019zero}, and Chen et al. \cite{chen2019data} train Generative Adversarial Networks (GANs) \cite{goodfellow2014generative} to generate training data for the entire KD process, utilizing the knowledge distilled from the teacher model. 
To free the limitation from a public dataset, a few research works consider data-free KD in FL.
In FedML \cite{shen2020federated}, latent knowledge from homogeneous models is applied to train heterogeneous models.
In FedHe \cite{fedhe2021}, logits belonging to the same class are directly averaged in a server. 
In FedGKT \cite{he2020group}, a neural network is split into a client and a server, while the server completes the entire training process based on the features and logits collected from all clients. Most of the existing data-free approaches are based on logits. 
While FedIN is also a data-free approach, instead of transmitting the logits, the knowledge exchanged in FedIN also includes the client features, which contain more information than simply using the logits.

\subsubsection{Sub-models}
To adapt to the available resources of different clients, several studies split the large models into small sub-models. HeteroFL \cite{diao2021heterofl} divides a large model into local models with different sizes. However, the architectures of local and global models are still restricted by the same model architecture. SlimFL \cite{baek2022joint} integrates slimmable neural network (SNN) architectures \cite{yu2019universally} into FL, adapting the widths of local neural networks based on resource limitations. In \cite{horvath2021fjord}, FjORD leverages Ordered Dropout and a self-distillation method to determine the model widths. However, similar to HeteroFL, they only vary the number of parameters for each layer. In this paper, we propose a more heterogeneous and flexible FL framework supporting various edge devices.

\begin{figure*}[!t]
\centering
\includegraphics[width=\textwidth]{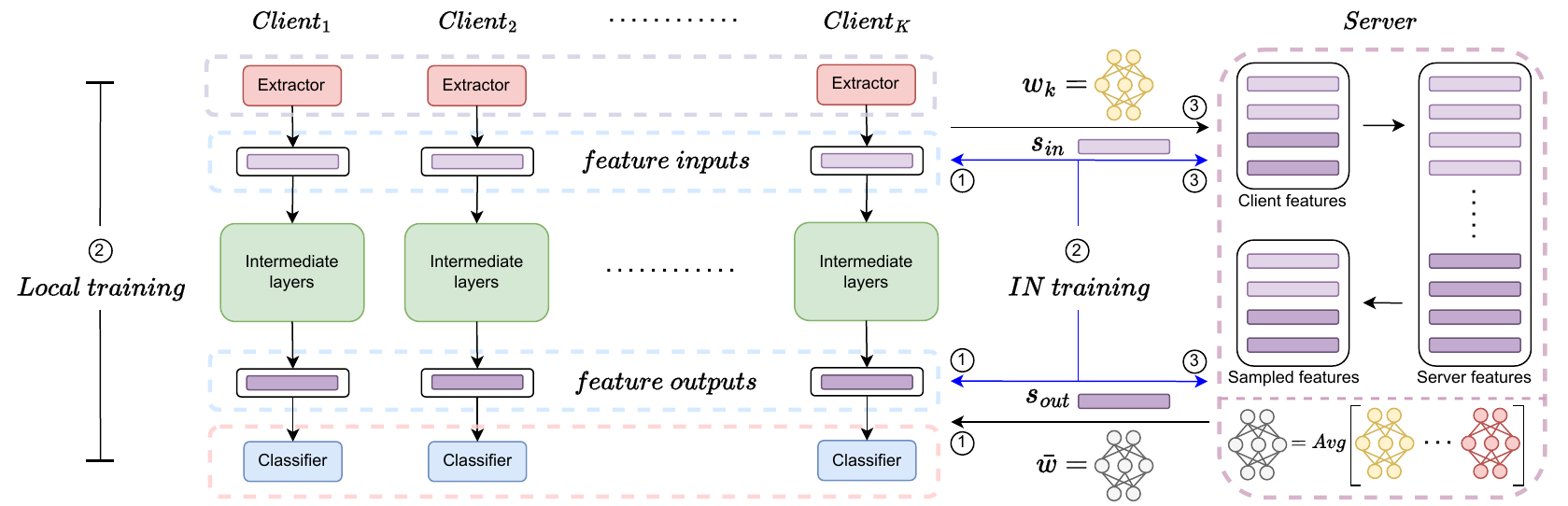}%
\caption{\textbf{Details of model architectures and the training process for FedIN.} In this figure, blue arrows represent the transmission of corresponding client features, i.e., feature inputs and feature outputs, $(s_{in}, s_{out})$. 
The process for FedIN is described as follows. {\normalsize{\textcircled{\footnotesize{1}}}} First, clients receive client features and global weights $\bar{w}$ from the server. {\normalsize{\textcircled{\footnotesize{2}}}} After updating client weights by global weights, the clients are training their models from the local private dataset and completing the IN training for the client features inputs and outputs $(s_{in},s_{out})$ from the server. {\normalsize\textcircled{\footnotesize{3}}} Upon completing the local training, clients transmit the model weights and new client features, denoted as $(w_k, s_{in}, s_{out})$, to the server. The aggregation methods for system heterogeneity are discussed in \sectionautorefname~\ref{sec:aggregation}.}
\label{fig_FedIN_process}
\end{figure*}

\section{Problem formulation}
We introduce the general federated learning problem for heterogeneous models in this section.

\subsection{Heterogeneous Federated Learning}
The goal of FL is to collaborate with the clients to train a shared global model while keeping their local data private. We briefly summarize the optimization problem below. 
We assume that $K$ clients participate in FL. 
Each client has a private dataset $D_k=\{(x_{i,k}, y_{i,k})|i=1,2,...,|D_k|\}$, where $k\in\{1,...,K\}$ is the index of a client, and $|D_k|$ denotes the size of a dataset $D_k$.
Private dataset $D_k$ is only accessible to client $k$, guaranteeing data privacy. 
In traditional FL, the clients share identical model architecture. We denote a training model by $f(x;w)$, where $w$ are the training weights and $x$ are the inputs.  
The loss function $l_k$ of client $k$ is shown as follows,
\begin{equation}
\label{FL_local_objective}
\begin{aligned}
 \min_w&\ & l_k(w)=\frac{1}{|D_k|}\sum_{i=1}^{|D_k|}{l(f(x_i;w), y_i)},
\end{aligned}
\end{equation}
where $l(\cdot,\cdot)$ is a loss function for each data sample $(x_{i,k}, y_{i,k})$.
If the total size of all datasets is $N=\sum_{i=1}^K |D_k|$, the global optimization target is,
\begin{equation}
\label{FL_global_objective}
\begin{aligned}
 \min_w&\ & L(w)=\sum_{k=1}^K\frac{|D_k|}{N}{l_k(w)},
\end{aligned}
\end{equation}
where $w$ are the parameters from the global shared model,
The size of a private dataset affects the weight of its loss in the global loss function. If a client has a larger dataset, its loss deserves more attention. A common method for updating $w$ in original federated learning is to aggregate the updated weights and gradients from different clients \cite{mcmahan2017communication}.

Nevertheless, it may not be possible to deploy an identical model architecture for all the clients due to system heterogeneity. One potential solution is to allow clients to select different model architectures according to their capabilities in heterogeneous FL. The problem of heterogeneous FL is described as follows. 
We denote $w_k$ as the model weights of client $k$.
Therefore, the global loss function is described as follows,
\begin{equation}
\label{Hetero_FL_global_objective}
\begin{aligned}
 \min_{w_1, w_2, ..., w_K}\ L(w_1, ..., w_K)=\sum_{k=1}^K\frac{|D_k|}{N}{l_k(w_k)},
\end{aligned}
\end{equation}
where the optimized model weights $\{w_1, w_2, ..., w_K\}$ have different size. Thus, the direct aggregation of entire model weights becomes unfeasible when dealing with heterogeneity among models. Therefore, we adopt layer-wise heterogeneous aggregation \cite{liu2022no} as an alternative approach to aggregate the layer weights of heterogeneous models instead of the entire model weights in our experiments.

\section{FedIN: Federated Intermediate Layers Learning}
In this section, we describe the details of FedIN, focusing on addressing system heterogeneity by deploying clients with diverse model architectures that align with their available resources.
\figurename~\ref{fig_FedIN_process} illustrates the workflow of FedIN. The client model consists of three key components: an extractor, intermediate layers, and a classifier.
The outputs of the extractor, referred to as feature inputs ($s_{in}$), serve as inputs to the intermediate layers. Similarly, the outputs of the intermediate layers, referred to as feature outputs ($s_{out}$), act as inputs to the classifier. The client features are the pair of feature inputs and outputs, denoted as $(s_{in}, s_{out})$.
To be specific, FedIN encompasses two training processes: local training, which leverages the private dataset, and IN training, which relies on the feature inputs and outputs $(s_{in}, s_{out})$. Moreover, to address the challenge of gradient divergence arising from conflicts between local training and IN training, we propose a convex optimization problem formulation to obtain the optimal updated gradients.

\subsection{Local Training and IN Training}
The clients receive a single batch of feature inputs and feature outputs, denoted as $S=\{(s_{i,in}^c, s_{i,out}^c)\}_{i=1}^{|S|}$, from the server. These samples are utilized for training the intermediate layers during the IN training process. The superscript $c$ means that these feature inputs and outputs are from the central server. The clients begin their local training after receiving a batch of client features from the server. For an instance $(x_{i,k},y_{i,k})\in D_k$, client $k$ conducts local training on its private dataset. The loss function of the local training is shown as follows,
\begin{equation}
\label{clients_local_loss}
\begin{aligned}
l_{local,k}&=&l_{CE}(f(x_{i,k};w_k^t), y_{i,k}) + \frac{\mu}{2}||w_k^t-w_k^{t-1}||^2,
\end{aligned}
\end{equation}
where $w_k^t$ are the weights of client $k$ at time $t$, and $l_{CE}$ is the cross-entropy loss function for the local training. To ensure client consistency and prevent overfitting, we add a proximal regularization term \cite{li2018federated} in Eq. \ref{clients_local_loss}.

The second training process is IN training, which is training the intermediate layers from the features dataset $S$. It is worth mentioning that the sample number of $S$ is one batch size. We denote the weights of the extractor and the classifier by $w_{e,k}$ and $w_{c,k}$ for client $k\in\{1,...,K\}$. Moreover, the weights of the intermediate layers are denoted by $w_{in,k}$. The relations among the data sample $(x_{i,k},y_{i,k})\in D_k$, client weights, and $(s_{i,in}^k, s_{i,out}^k)$ are shown as follows,
\begin{equation}
\label{client_sin}
\begin{aligned}
s_{i,in}^k=f(x_{i,k};w_{e,k}),
\end{aligned}
\end{equation}
\begin{equation}
\label{client_sout}
\begin{aligned}
s_{i,out}^k=f(s_{i,in}^k;w_{in,k}),
\end{aligned}
\end{equation}
\begin{equation}
\label{client_sout_equal}
\begin{aligned}
f(x_{i,k};w_k)=f(s_{i,out}^k;w_{c,k}).
\end{aligned}
\end{equation}
Eq. \ref{client_sin} shows that the feature input $s_{i,in}^k$ is the output of the extractor $w_{e,k}$ of an instance $(x_{i,k},y_{i,k})$ from client $k$. Eq. \ref{client_sout} describes that the feature output $s_{i,out}^k$ is the output of the intermediate layers $w_{in,k}$ with the feature input $s_{i,in}^k$. Eq. \ref{client_sout_equal} proves the equivalence between the output of the classifier $w_{c,k}$ and the output of the whole client model $w_k$. In the local training process, the feature inputs and outputs $(s_{i,in}^k,s_{i,out}^k)$ are collected by client $k$. After completing the training process, client $k$ transmits one batch of the collected feature inputs and outputs to the server. This process is indicated by the blue arrows in \figurename~\ref{fig_FedIN_process}.

Eq. \ref{client_sout} shows the main function of the IN training, as shown in \figurename~\ref{fig_FedIN_process}. After the client receives the feature dataset $S=\{(s_{i,in}^c, s_{i,out}^c)\}_{i=1}^{|S|}$, it begins the IN training for the intermediate layers. 
The feature inputs $s_{i,in}^c$ from the server are the inputs of the intermediate layers, while the $s_{i,out}^c$ are the targets of the IN training. The loss function of IN training is defined as follows,
\begin{equation}
\label{clients_IN_loss}
\begin{aligned}
l_{IN,k}=l_{MSE}(f(s_{i,in}^c;w_{in,k}), s_{i,out}^c),
\end{aligned}
\end{equation}
where $l_{MSE}$ is a mean-square error loss function. The weights $w_{in,k}$ are updated by the loss functions of the local training $l_{local,k}$ and the IN training $l_{IN,k}$.

\subsection{Gradient Alleviation} \label{Gradient_divergence}
However, it is possible that the gradients from the local training and the IN training are divergent, which would drag the convergent speed and disturb the model to achieve the optimum point \cite{wang2020tackling, zhao2018federated}. It is crucial to alleviate the gradients divergence problem in our method.
Therefore, how to mitigate this problem is critical in FedIN. To address this problem, we formulate a convex optimization problem as follows. 

We define the gradients from the local training as a matrix $G_{local}$ and the gradients from the IN training as a matrix $G_{IN}$. To guarantee the optimized direction of the models, we design a constraint for the gradient as follows,
\begin{equation}
\label{gradient_constraint}
\begin{aligned}
\langle G_{IN},G_{local} \rangle \geq 0,
\end{aligned}
\end{equation}
where $\langle \cdot,\cdot \rangle$ is the dot product, which ensures the optimized direction for $G_{local}$ and $G_{IN}$ to be the same. In the optimization problem, we denote the new optimized gradients by a matrix $Z$ and model the following convex optimization primal problem,
\begin{equation}
\label{primal_optim}
\begin{aligned}
 \min_Z && ||G_{IN}-Z||_{F}^{2}, \\
 s.t.\ && \langle Z,G_{local}\rangle \geq 0,
\end{aligned}
\end{equation}
where we maintain the optimized direction between $Z$ and $G_{local}$ to be the same and minimize the distance between $Z$ and $G_{in}$. We consider that the information from the feature inputs and outputs is more fruitful than the local private dataset which is easier to have over-fitting in the training process. We solve this convex optimization problem by the Lagrange dual problem \cite{bot2009duality}. The Lagrangian is shown as,
\begin{equation}
\label{Lagrangian}
\begin{aligned}
 L(Z,\lambda)=tr(G_{IN}^TG_{IN})-tr(Z^TG_{IN})\\
 -tr(G_{IN}^TZ)+tr(Z^TZ)-\lambda tr(G_{local}^TZ),
\end{aligned}
\end{equation}
where $tr(A)$ means the trace of the matrix $A$, and the $\lambda$ is a Lagrange multiplier associated with $\langle Z,G_{local}\rangle \geq 0$. To derive the dual problem, we first get the optimum of $Z$ for the Lagrangian Eq. \ref{Lagrangian}, and then obtain the Lagrange dual function $g(\lambda)=\inf_Z L(Z,\lambda)$. Thus, the Lagrange dual problem is described as follows,
\begin{equation}
\label{dual_optim}
\begin{aligned}
 \max_\lambda\ &g(\lambda)=-\frac{\lambda^2}{4}tr(G_{local}^TG_{local})\\
 &-\lambda tr(G_{local}^TG_{IN}), \\
 s.t.\ &\lambda \geq 0,
\end{aligned}
\end{equation}
where the optimum of the Lagrangian Eq. \ref{Lagrangian} is $Z=G_{IN}+\frac{\lambda}{2}G_{local}$. If the $\lambda$ is large enough, it is obvious that $\langle Z,G_{local}\rangle > 0$, which means this convex optimization problem holds strong duality because it satisfies the Slater's constraint qualification\cite{boyd2004convex}, i.e., the optimum of the primal problem Eq. \ref{primal_optim} is also $Z=G_{IN}+\frac{\lambda}{2}G_{local}$. Furthermore, the dual problem Eq. \ref{dual_optim} can be solved to obtain the analytic solution for  $\lambda$ and $Z$, which is shown as follows,
\begin{equation}
\label{analytic_sol}
    Z=
\begin{cases}
G_{IN},& \text{if } b\geq 0 \\
G_{IN}-\frac{b}{a}G_{local}, & \text{if } b<0
\end{cases}
\end{equation}
where $a=tr(G_{local}^T G_{local})$ and $b=tr(G_{local}^T G_{IN})$. However, one crucial point is that the clients will handle this optimization process. If we calculate each gradient matrix following Eq. \ref{analytic_sol}, this process would occupy lots of computing resources because of the matrix multiplication. Therefore, to mitigate the computational pressure on the clients, we simplified the updated gradient matrix as,
\begin{equation}
\label{updated_gradients}
\begin{aligned}
Z=G_{IN}+\frac{\lambda}{2}G_{local},
\end{aligned}
\end{equation}
where $\lambda=1$ is set for the optimum point of the primal problem in our experiment settings. Since $G_{IN}$ is only associated with the weights $w_{IN,k}$ and not related to $w_{e,k}$ and $w_{c,k}$, the client models are optimized by Eq. \ref{updated_gradients} in FedIN directly.

\subsection{Weight Aggregation}
\label{sec:aggregation}
FedIN does not impose any limitations on the model architecture, indicating that diverse model architectures can be deployed in FedIN by leveraging a single batch of client features as the communication knowledge. If client models have different numbers of layers, FedIN adopts layer-wise heterogeneous aggregation \cite{liu2022no}, enabling the server to aggregate weights from the same layer rather than the same model. Similarly, when client models have different architectures, FedIN aggregates model weights only from models with identical architectures, the same as the aggregation method used in FedAvg \cite{mcmahan2017communication} and FedDF \cite{lin2020ensemble}. For example, the weights of CNNs cannot be aggregated with the weights of Transformers \cite{vaswani2017attention}. However, the weights of CNNs can be aggregated with other CNNs possessing the same depths and widths.
The effectiveness of FedIN with these two distinctive aggregation methods is further demonstrated in our experimental section.

\subsection{Detailed Algorithm of FedIN}
The algorithm process of FedIN is presented in Algorithm \ref{Algorithm_FedIN}. On the server side, the server receives the model weights, feature inputs and outputs $(w_k, s_{i,in}^k, s_{i,out}^k)$. The new feature inputs and outputs, $(s_{i,in}^k, s_{i,out}^k)$, are stored in the server dataset. The server samples one batch of feature inputs and outputs, denoted as $S=\{(s_{i,in}^c, s_{i,out}^c)\}_{i=1}^{|S|}$, as indicated in line 5 of Algorithm \ref{Algorithm_FedIN}. Finally, the server transmits averaged weights $\bar{w}$ and the batch of feature inputs and outputs $S$ to all the clients. On the client side, no data is received if it is in the initial training process. Each client $k$ performs local training using Eq. \ref{clients_local_loss}. During the initial training, client $k$ also collects the feature inputs and outputs, $(s_{i,in}^k, s_{i,out}^k)$. If client $k$ is not in the initial training, it receives $(\bar{w}, S)$ from the server. Client $k$ computes $l_{local,k}$ from the local private dataset and $l_{IN,k}$ from the feature dataset $S$, and updates its model by Eq. \ref{updated_gradients}. At last, client $k$ transmits $(w^k, s_{i,in}^k, s_{i,out}^k)$ to the server.

\begin{algorithm}[htbp]
    \caption{FedIN}
    \label{Algorithm_FedIN}
    \KwIn {Local dataset $D_k, k\in\{1,...,K\}$, $K$ clients and their weights $w_1, ..., w_K$.}
    \KwOut {Optimal weights for all the clients $w_1, ..., w_K$.}
    \textbf{Server process:}\\
    \While {not converge}
    {   
        Receives $(w_k, s_{i,in}^k, s_{i,out}^k)$ from the client $k$.\\
        Saves feature inputs and outputs $(s_{i,in}^k, s_{i,out}^k)$ in the server dataset.\\
        Randomly samples the one batch of feature inputs and outputs $S=\{(s_{i,in}^c, s_{i,out}^c)\}_{i=1}^{|S|}$ from the server dataset.\\
        Transmits averaged weights and a batch of feature inputs and outputs $(\bar{w}, S)$ to the sampled clients.
    }
    \textbf{Client processes:} \\
    \While {random sampled clients $k, k\in {1,...,K}$}
    {
        \If {initial training} 
        {
            Updates the client $k$ model according to the local training Eq. \ref{clients_local_loss}.\\
            Collects $(s_{i,in}^k, s_{i,out}^k)$ from data samples $(x_{i,k},y_{i,k})\in D_k$ from the local training. \\
        }
        \Else
        {
            Receives $(\bar{w}, S)$ from the server.\\
            $w_k = \bar{w}$. \\
            Computes $l_{local,k}$ based on the private dataset by Eq. \ref{clients_local_loss}. \\
            Collects $(s_{i,in}^k, s_{i,out}^k)$ from data samples $(x_{i,k},y_{i,k})\in D_k$ from the local training. \\
            Computes $l_{IN,k}$ according to the $S$ by Eq. \ref{clients_IN_loss}. \\
            Updates the weights by Eq. \ref{updated_gradients}.
        }
        Transmit the weights and collected information $(w_k, s_{i,in}^k, s_{i,out}^k)$ to the server.
    }
\end{algorithm}

\section{Experiments}
In this section, we conduct experiments to evaluate the performances of FedIN on the CIFAR-10 \cite{krizhevsky2009learning}, Fashion-MNIST \cite{xiao2017/online} and SVHN \cite{netzer2011reading} datasets, and also manage the ablation studies of FedIN to assess the effectiveness of individual components within FedIN. Our codes will be released on Github.

\subsection{Experiment Settings}
\subsubsection{Federated settings}
\begin{figure}[!t]
\centering
\subfloat[IID data.]
{\includegraphics[width=4cm]{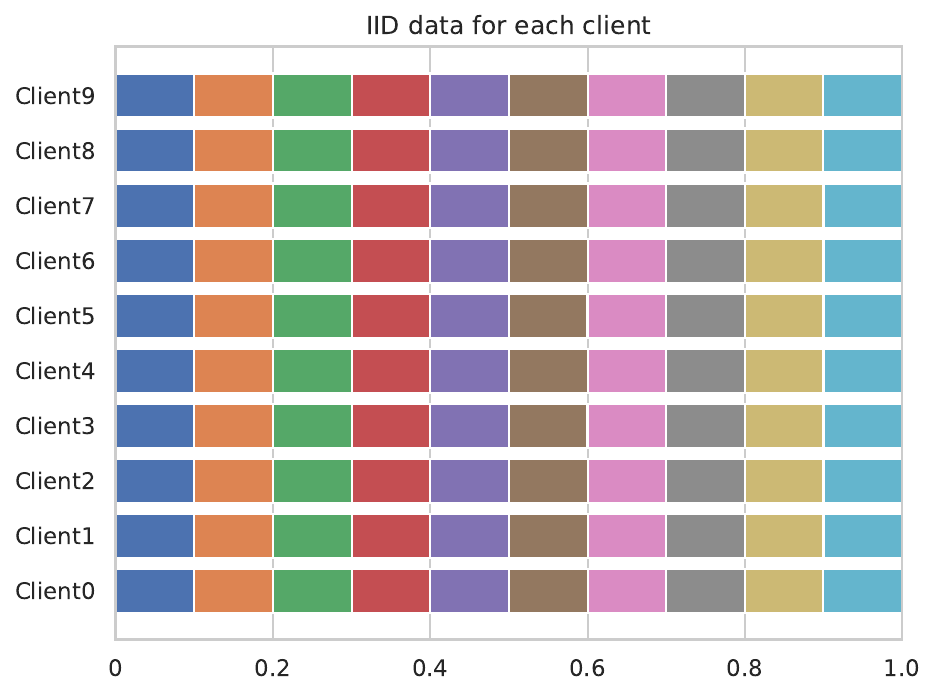}%
\label{fig_IID}}
\hfil
\subfloat[Non-IID data.]
{\includegraphics[width=4cm]{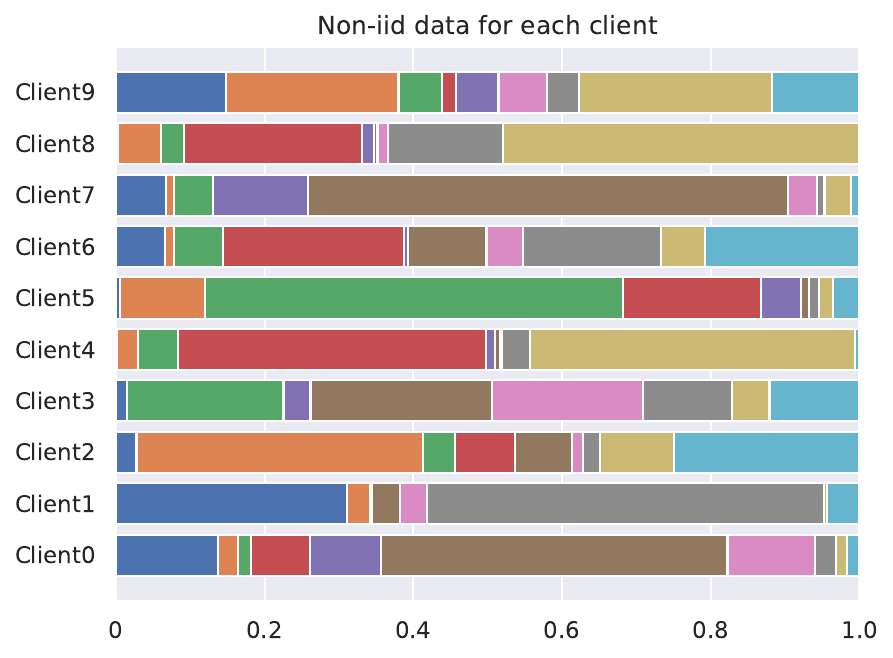}%
\label{fig_non_IID}}
\caption{Illustrations for IID data and non-IID data with $\alpha=0.5$.}
\label{fig_data_distribution}
\end{figure}

CIFAR-10 contains 50,000 $32\times 32$ color images in the training dataset and 10,000 images in the testing dataset with ten different classes, such as car, dog, and cat. 
Fashion-MNIST has 70,000 $28\times 28$ gray-scale images with ten classes, such as T-shirts, trousers, and dresses, including 60,000 samples for training and 10,000 samples for validation. 
SVHN is a real-world image dataset, obtained from house numbers in Google Street View images. It includes 73,257 $32\times 32$ color images for training and 26,032 images for testing.
We establish two distributions for these datasets, independent and identically distribution (IID), and non-IID, as demonstrated in \figurename~\ref{fig_data_distribution}. The non-IID data is generated using a Dirichlet distribution with a parameter $\alpha=0.5$, as elaborated in \figurename~\ref{fig_non_IID}. 
We have 100 clients in the FL training process. The model architectures are ResNet10, ResNet14, ResNet18, ResNet22 and ResNet26 from PyTorch source codes, and evenly distributed among 100 clients. 
The number of communication rounds is set to 500. The batch size is 16 during the training process.
For all datasets, the clients complete 5 epochs of inner training during each communication round.

\subsubsection{Baselines} We have two classic algorithms, FedAvg and FedProx, and five state-of-the-art methods, Scaffold, FedNova, MOON, HeteroFL, and InclusiveFL, as our baselines.
\begin{itemize}
    \item \textbf{FedAvg} \cite{mcmahan2017communication}: Clients transmit their updated gradients to the server, and the server performs gradient averaging to update the global model.
    \item \textbf{FedProx} \cite{li2018federated}: Based on FedAvg, clients incorporate a regularization term to minimize the disparities between local models and a global model.
    \item \textbf{Scaffold} \cite{karimireddy2020scaffold}: Clients retain the local control variate, while the server maintains server control variate, effectively mitigating the impact of client-drift.
    \item \textbf{FedNova} \cite{wang2020tackling}: The server uses a normalized averaging method that mitigates objective inconsistency while maintaining fast error convergence.
    \item \textbf{MOON} \cite{li2021model}: Clients conduct the contrastive learning between the current local model, global model, and the local model from the previous time stamp.
    \item \textbf{HeteroFL} \cite{diao2021heterofl}: The smaller models are derived from the largest model sharing the same architecture. These small models are deployed on the clients. The corresponding parameters from all models are aggregated to update the largest model.
    \item \textbf{InclusiveFL} \cite{liu2022no}: This method proposes a momentum knowledge distillation method to enhance the transfer of knowledge from large models to smaller models.
\end{itemize}

\subsubsection{Details of baselines}
FedIN and the baselines, with the exception of HeteroFL, utilize the layer-wise aggregation technique proposed in \cite{liu2022no} under our heterogeneous model environment. Since HeteroFL requires model splitting based on its own methodology, it cannot utilize this aggregation technique. Therefore, in order to maintain a similar number of parameters as the other baselines, we deploy ResNet152 in HeteroFL instead of using the largest model, ResNet26, as in other methods. The model split mode in HeteroFL is "dynamic\_a1-b1-c1-d1-e1" from the source code because of five heterogeneous models in all other methods.
The hyper-parameter $\frac{\mu}{2}$ for FedProx and FedIN is 0.05.
We use Adam optimizer with a learning rate of 0.001, $\beta_1=0.9$ and $\beta_2=0.999$, default parameter settings for all methods.
All experiments are conducted in the same environment utilizing four Nvidia RTX3090 GPUs.

\newcommand{\mycc}{\cellcolor{Gray}}

\begin{table}[!t]
    \def\arraystretch{1.6}
    \caption{Model accuracy for IID and non-IID data of FashionMNIST. Target accuracy is 85.}
    \label{tab:acc_FashionMNIST}
    \centering
    \begin{tabular}{ccccc}
    \hline
        \multirow{2}*{Methods} & \multicolumn{4}{c}{FashionMNIST (ACC=85)} \\
        \cline{2-5}
        & IID $\uparrow$ & Non-IID $\uparrow$ & Round $\downarrow$ & Speedup $\uparrow$  \\
      \hline
       
        FedAvg\cite{mcmahan2017communication} & 90.3 & 89.4 & 47 & $\times$ 1.0 \\

        FedProx\cite{li2018federated} & 89.7 & 87.6 & 40 & $\times$ 1.2\\

        Scaffold\cite{karimireddy2020scaffold} & 88.3 & 87.1 & 25 & $\times$ 1.9\\

        FedNova\cite{wang2020tackling} & 87.5 & 87.3 & 36 & $\times$ 1.3 \\

        MOON\cite{li2021model} & 89.5 & 89.0 & 34 & $\times$ 1.4\\
       
        HeteroFL\cite{diao2021heterofl} & 89.3 & 89.5 & 140 & $\times$ 0.3 \\

        InclusiveFL\cite{liu2022no} & 88.4 & 89.1 & 31 & $\times$ 1.5\\

        \mycc FedIN (Ours) & \mycc \textbf{91.2} & \mycc \textbf{90.3} & \mycc \textbf{20} & \mycc \textbf{$\times$ 2.4} \\

    \hline
    \end{tabular}
\end{table}

\begin{table}[!t]
    \def\arraystretch{1.6}
    \caption{Model accuracy for IID and non-IID data of SVHN. Target accuracy is 80.}
    \label{tab:acc_SVHN}
    \centering
    \begin{tabular}{ccccc}
    \hline
        \multirow{2}*{Methods} & \multicolumn{4}{c}{SVHN (ACC=80)} \\
        \cline{2-5}
        & IID $\uparrow$ & Non-IID $\uparrow$ & Round $\downarrow$ & Speedup $\uparrow$  \\
      \hline
       
        FedAvg\cite{mcmahan2017communication} & 89.2 & 84.5 & 82 & $\times$ 1.0 \\

        FedProx\cite{li2018federated} & 90.6 & 87.3 & 45 & $\times$ 1.8\\

        Scaffold\cite{karimireddy2020scaffold} & 91.1 & 86.0 & 72 & $\times$ 1.1\\

        FedNova\cite{wang2020tackling} & 87.3 & 86.7 & 106 & $\times$ 0.8 \\

        MOON\cite{li2021model} & 89.5 & 86.1 & 55 & $\times$ 1.5\\
       
        HeteroFL\cite{diao2021heterofl} & \textbf{93.8} & 89.3 & 107 & $\times$ 0.8 \\

        InclusiveFL\cite{liu2022no} & 90.9 & 88.7 & 67 & $\times$ 1.2\\

        \mycc FedIN (Ours) & \mycc \textbf{91.8} & \mycc \textbf{89.3} & \mycc \textbf{29} & \mycc \textbf{$\times$ 2.8} \\

    \hline
    \end{tabular}
\end{table}

\begin{table}[!t]
    \def\arraystretch{1.6}
    \caption{Model accuracy for IID and non-IID data of CIFAR-10. Target accuracy is 60.}
    \label{tab:acc_CIFAR10}
    \centering
    \begin{tabular}{ccccc}
    \hline
        \multirow{2}*{Methods} & \multicolumn{4}{c}{CIFAR-10 (ACC=60)} \\
        \cline{2-5}
        & IID $\uparrow$ & Non-IID $\uparrow$ & Round $\downarrow$ & Speedup $\uparrow$  \\
      \hline
       
        FedAvg\cite{mcmahan2017communication} & 76.8 & 66.2 & 109 & $\times$ 1.0 \\

        FedProx\cite{li2018federated} & 77.6 & 72.0 & 72 & $\times$ 1.5\\

        Scaffold\cite{karimireddy2020scaffold} & 79.0 & 68.1 & 120 & $\times$ 0.9\\

        FedNova\cite{wang2020tackling} & 62.9 & 60.3 & 229 & $\times$ 0.5 \\

        MOON\cite{li2021model} &  74.1 & 67.4 & 129 & $\times$ 0.8\\
       
        HeteroFL\cite{diao2021heterofl} & 72.1 & 61.0 & 273 & $\times$ 0.4 \\

        InclusiveFL\cite{liu2022no} & 75.0 & 66.1 & 160 & $\times$ 0.7\\

        \mycc FedIN (Ours) & \mycc \textbf{80.5} & \mycc \textbf{75.9} & \mycc \textbf{54} & \mycc \textbf{$\times$ 2.0} \\

    \hline
    \end{tabular}
\end{table}

\begin{figure}[!t]
\centering
\includegraphics[width=8.3cm]{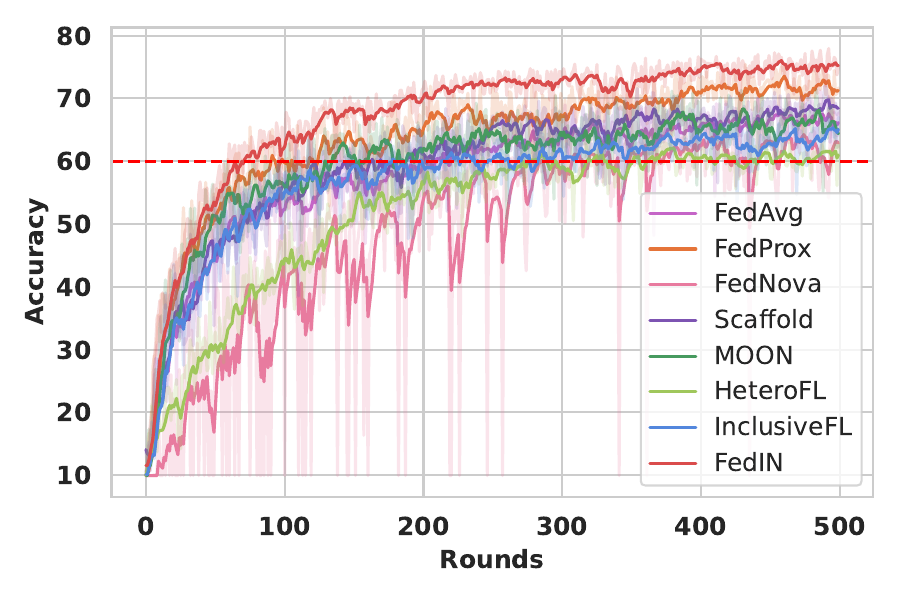}%
\caption{The smoothed test accuracy on non-IID data of CIFAR-10. The original results of accuracy are the grey lines. The red dot line denotes the target accuracy in \tablename~\ref{tab:acc_CIFAR10}.}
\label{fig:acc_noniid_cifar10}
\end{figure}

\subsection{Accuracy Analyses.}

\subsubsection{Accuracy of IID and non-IID data.}
We conduct experiments on the IID and non-IID data in Fashion-MNIST, SVHN, and CIFAR-10 datasets. The experiment results are shown in \tablename~\ref{tab:acc_FashionMNIST}, \tablename~\ref{tab:acc_SVHN}, and \tablename~\ref{tab:acc_CIFAR10}.  These tables provide details on the communication round (denoted as "Round") at which the methods achieve the target accuracy (ACC) under the non-IID setting. The best results in each table are highlighted in bold. The symbols "$\uparrow$" and "$\downarrow$" indicate that a higher or lower value of the respective metric is better, respectively. The target accuracy for non-IID data is specified at the top of each table.

From \tablename~\ref{tab:acc_FashionMNIST}, FedIN achieves the highest accuracy among all methods. It attains an accuracy of 91.2\% on IID data and 90.3\% on non-IID data. Furthermore, FedIN requires only 20 communication rounds to reach the target accuracy, demonstrating a speedup of 2.4 times compared to the baseline FedAvg.

In \tablename~\ref{tab:acc_SVHN}, FedIN achieves the best results with 91.8\% on IID and 89.3\% on non-IID data. FedAvg requires 82 communication rounds to achieve the target accuracy of SVHN (80\%), while FedIN accomplishes it in just 29 rounds. FedIN exhibits the fastest convergence speed among all state-of-the-art baselines, with a speedup of 2.8 times compared to FedAvg.

In \tablename~\ref{tab:acc_CIFAR10}, FedIN demonstrates substantial improvements compared to most baselines. It achieves an accuracy of 80.5\% on IID data and 75.6\% on non-IID data, surpassing the second-best baseline, FedProx, which achieves 77.6\% and 72.0\%, respectively. Notably, two baselines, HeteroFL \cite{diao2021heterofl} and InclusiveFL \cite{liu2022no}, which focus on addressing system heterogeneity, perform worse than FedIN, especially on non-IID data. HeteroFL achieves 72.1\% on IID data and 61.0\% on non-IID data, while InclusiveFL attains 75\% on IID and 66.1\% on non-IID data.
Moreover, the target accuracy for CIFAR-10 is 60\%, and FedAvg requires 109 rounds to achieve it, whereas FedIN only needs 54 rounds. HeteroFL requires 273 rounds and InclusiveFL needs 160 rounds, indicating slower convergence speeds compared to FedAvg. However, FedIN achieves a speedup of 2.0 times compared to FedAvg, demonstrating significantly accelerated convergence.

Additionally, \figurename~\ref{fig:acc_noniid_cifar10} shows the smoothed test accuracy on non-IID data of CIFAR-10.
FedIN (red line) achieves the highest accuracy and exhibits the fastest convergence speed throughout the training process. It is the first method to achieve the target accuracy (red dot line). 
FedProx\cite{li2018federated} is the second-best method but still lags behind FedIN. Several methods, including FedAvg\cite{mcmahan2017communication}, Scaffold\cite{karimireddy2020scaffold}, MOON\cite{li2021model}, and InclusiveFL\cite{liu2022no}, demonstrate similar convergence processes, achieving similar results ranging from 64\% to 68\%, as indicated in \tablename~\ref{tab:acc_CIFAR10}.
FedNova\cite{wang2020tackling} and HeteroFL\cite{diao2021heterofl} have significantly slower convergence speeds compared to FedAvg in \figurename~\ref{fig:acc_noniid_cifar10}, achieving only 60.3\% and 61.0\%, respectively. Moreover, FedIN incurs only a small additional overhead of one batch of feature inputs and outputs compared to FedAvg, as shown in \tablename~\ref{tab:overheads}.

\begin{table}[!t]
    \def\arraystretch{1.6}
    \caption{Model accuracy with homogeneous models.}
    \label{tab:acc_CIFAR10_homogeneous_model}
    \centering
    \begin{tabular}{cccccc}
    \hline
        \multirow{2}*{CIFAR-10} & \multicolumn{5}{c}{Methods} \\
        \cline{2-6}
        & FedProx & Scaffold & FedNova & MOON & \mycc FedIN  \\
      \hline
       
        IID & 83.5 & 84.3 & 82.0 & 84.2 &\mycc \textbf{84.7} \\

        Non-IID & 77.5 & 76.8 & 75.4 & 78.2 & \mycc \textbf{79.2}\\

    \hline
    \end{tabular}
\end{table}

\subsubsection{Accuracy of homogeneous models}
While FedIN primarily addresses the system heterogeneity challenge in FL, we also conduct experiments in a homogeneous model environment using CIFAR-10. All client models are ResNet18 in this experiment and the remaining federated settings are the same as the system heterogeneity experiments. As presented in \tablename~\ref{tab:acc_CIFAR10_homogeneous_model}, FedIN still outperforms state-of-the-art baselines, specifically designed to enhance FL performance in homogeneous model environments. Notably, FedIN achieves the highest accuracy, 84.7\% on IID data and 79.2\% on non-IID data of CIFAR-10, while the second-best result is 84.3\% from Scaffold on IID data and 78.2\% from MOON on non-IID data.

\begin{table*}[!t]
    \def\arraystretch{1.6}
    \small
    \caption{Model accuracy with heterogeneity models with FedAvg aggregations.}
    \label{tab:acc_FashionMNIST_not_agg}
    \centering
    \begin{tabular}{cccccccc}
    \hline
        \multirow{2}*{Fashion-MNIST} & \multicolumn{7}{c}{Methods} \\
        \cline{2-8}
        & FedAvg & FedProx & Scaffold & FedNova & MOON & InclusiveFL & \mycc FedIN  \\
      \hline
       
        IID & 86.1 & 83.4 & 87.7 & 84.2 & 87.0 & 88.1 & \mycc \textbf{88.9} \\

        Non-IID & 85.4 & 82.1 & 86.3 & 83.9 & 86.5 & 86.4 & \mycc \textbf{88.0} \\

    \hline
    \end{tabular}
\end{table*}

\subsubsection{Accuracy with FedAvg aggregation}
To ensure a fair comparison, both the baselines and FedIN employ layer-wise aggregation. However, it is worth noting that FedIN can be deployed in scenarios with extreme heterogeneity, where layer-wise aggregation is not feasible. In such cases, model weights with the same architectures are the only ones that can be aggregated. To demonstrate the effectiveness of FedIN in such extreme environments, we conducted experiments on the Fashion-MNIST dataset, utilizing FedAvg aggregation. The remaining federated settings are the same in this experiment.
As indicated in \tablename~\ref{tab:acc_FashionMNIST_not_agg}, FedIN still achieves the highest accuracy, 88.9\% on IID data and 88.0\% on non-IID data. These results further emphasize the effectiveness of FedIN in extreme system heterogeneity environments.

\begin{table}[!t]
    \def\arraystretch{1.6}
    \scriptsize
    \caption{Training overheads for different methods. "Params" indicates the communication overheads. "Memory" refers to the memory occupied by methods in the training process.}
    \label{tab:overheads}
    \centering
    \begin{tabular}{cccccc}
    \hline
        \multirow{2}*{Metrics} & \multicolumn{5}{c}{Methods} \\
        \cline{2-6}
         & FedAvg & Scaffold & MOON & HeteroFL &  \mycc FedIN  \\
      \hline
       
        Params(M) $\downarrow$ & 12.28 & 24.56 & 12.28 & 16.29  & \mycc \textbf{12.35} \\

        Memory(MB) $\downarrow$ & 235.0 & 470.0 & 705.0 & 445.6  & \mycc \textbf{235.3} \\

    \hline
    \end{tabular}
\end{table}

\begin{figure}[!t]
\centering
\subfloat[CKA similarity for IID data in CIFAR-10.]
{\includegraphics[width=4cm]{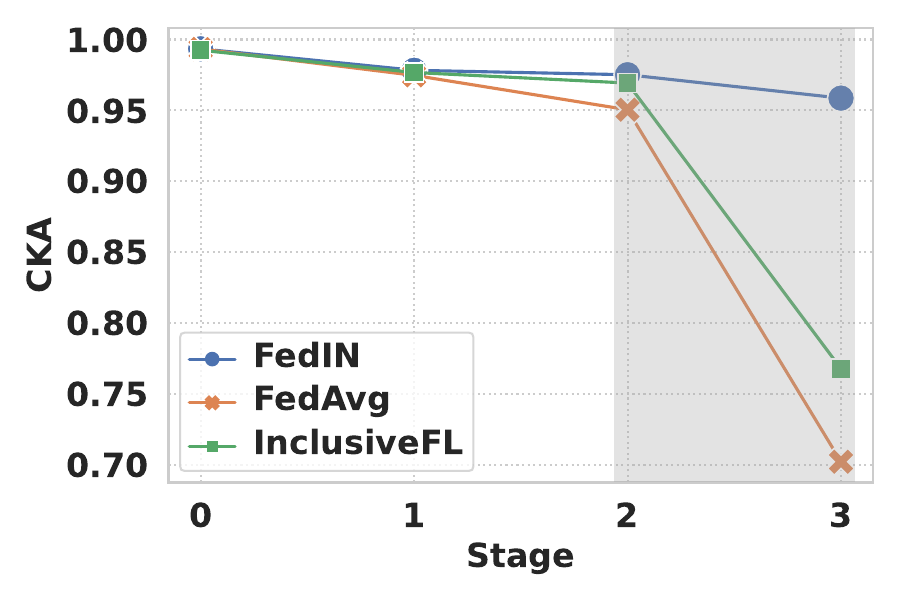}%
\label{fig_CKA_iid}}
\hfil
\subfloat[CKA similarity for non-IID data in CIFAR-10.]
{\includegraphics[width=4cm]{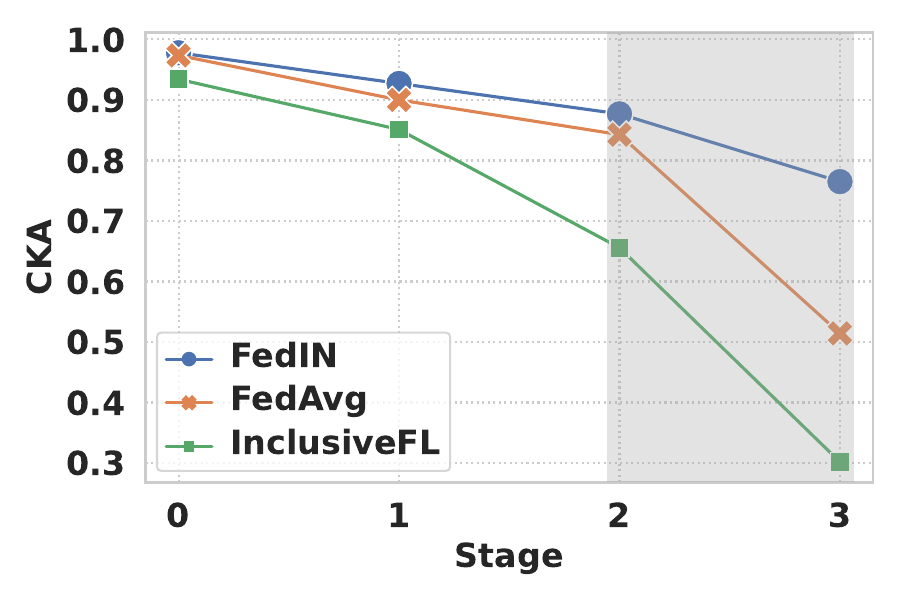}%
\label{fig_CKA_non_iid}}
\caption{Illustrations for CKA similarity of IID data and non-IID data with CIFAR-10.}
\label{fig_CKA}
\end{figure}

\begin{figure}[!t]
\centering
\subfloat[Stage 2 of FedAvg.]
{\includegraphics[width=4cm]{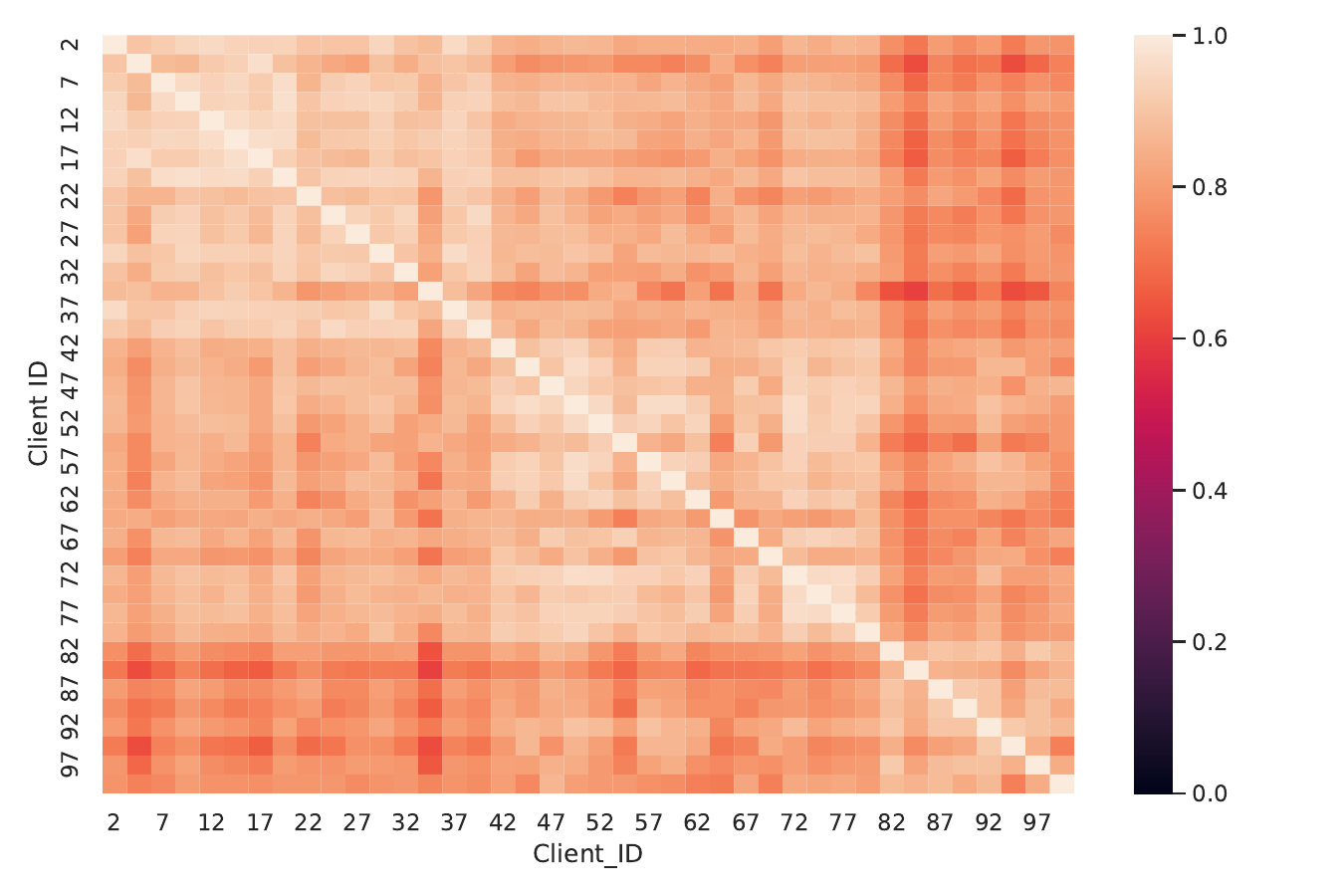}%
\label{fig:cka_fedavg_s2}}
\hfil
\subfloat[Stage 3 of FedAvg.]
{\includegraphics[width=4cm]{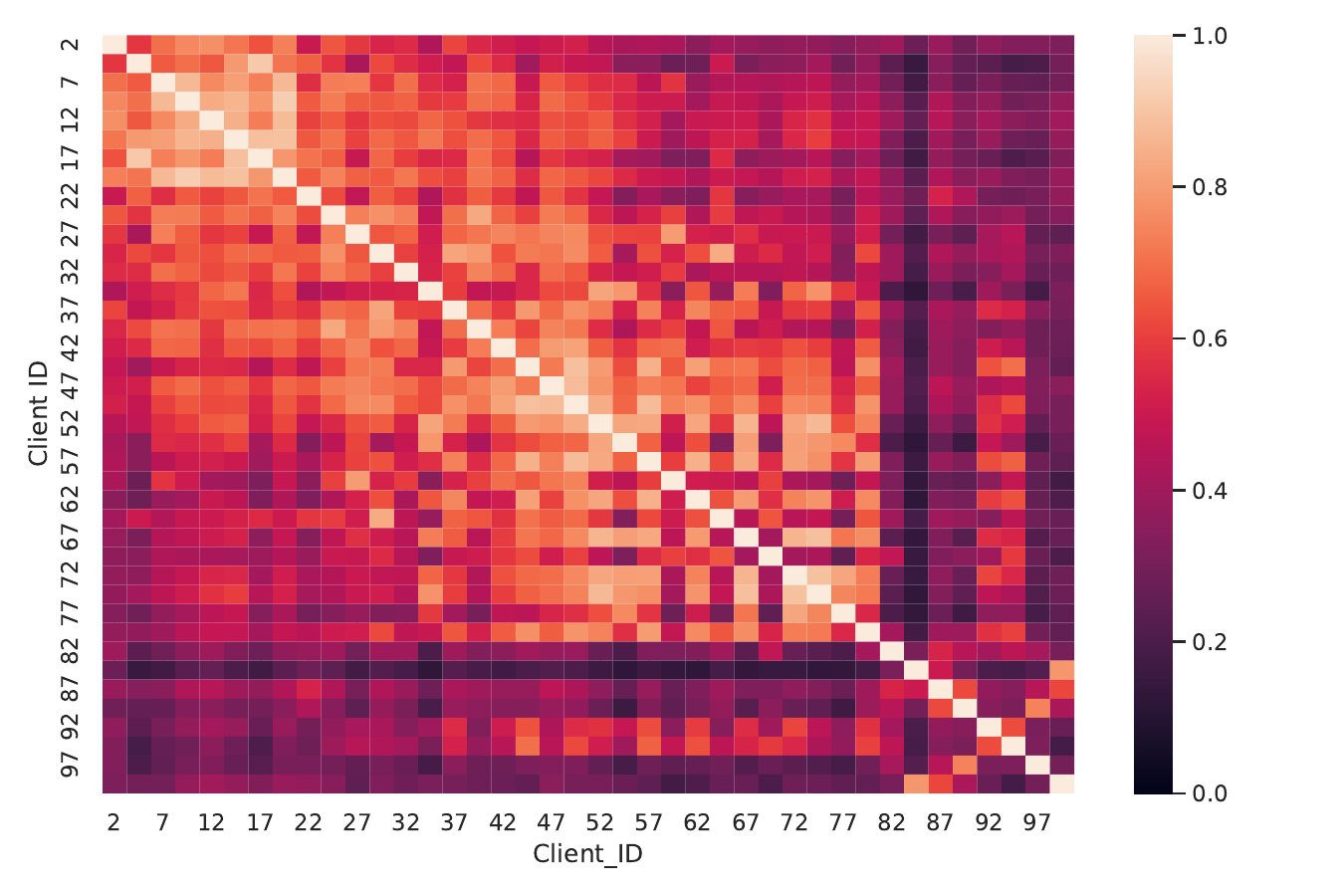}%
\label{fig:cka_fedavg_s3}}

\subfloat[Stage 2 of InclusiveFL.]
{\includegraphics[width=4cm]{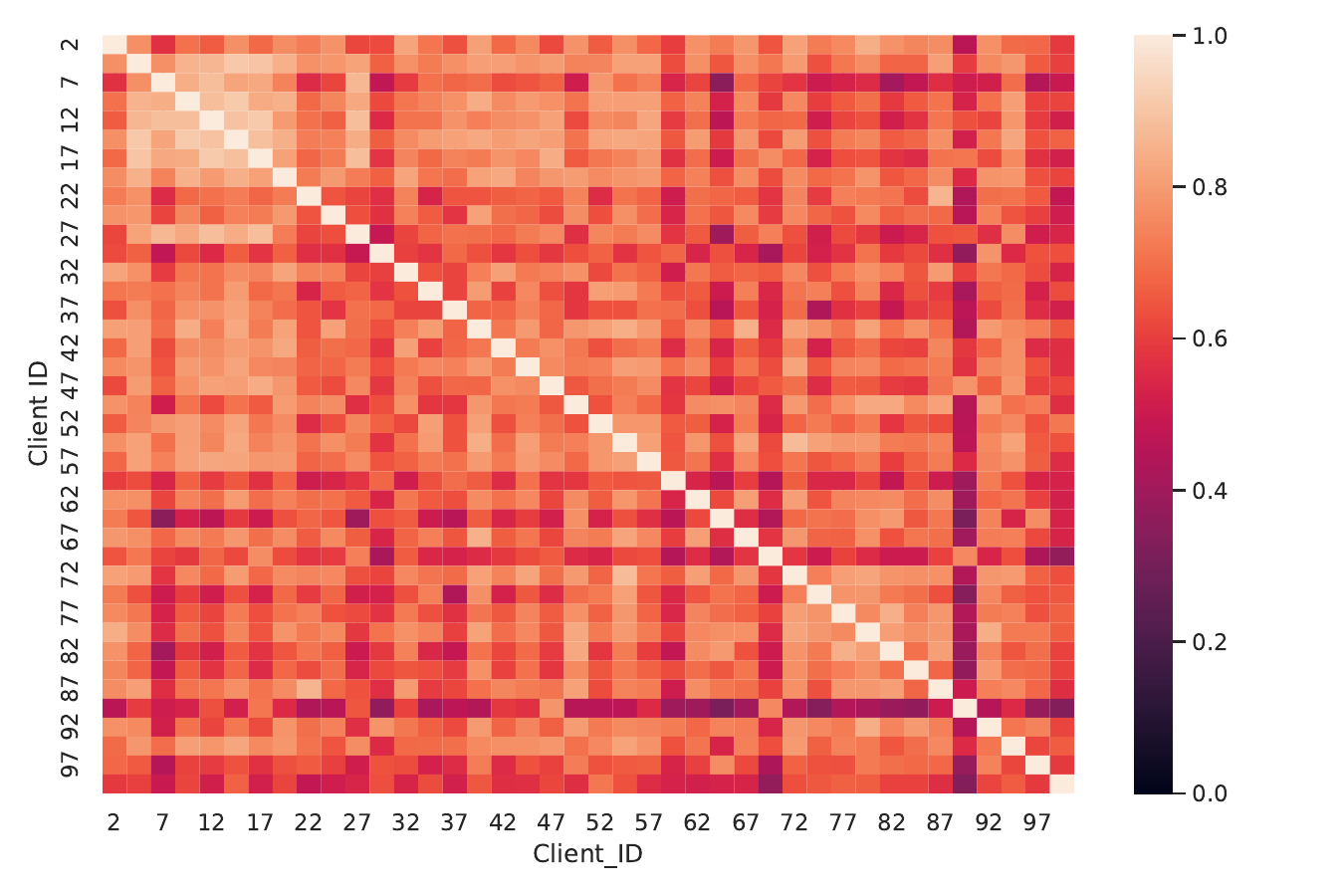}%
\label{fig:cka_inclusivefl_s2}}
\hfil
\subfloat[Stage 3 of InclusiveFL.]
{\includegraphics[width=4cm]{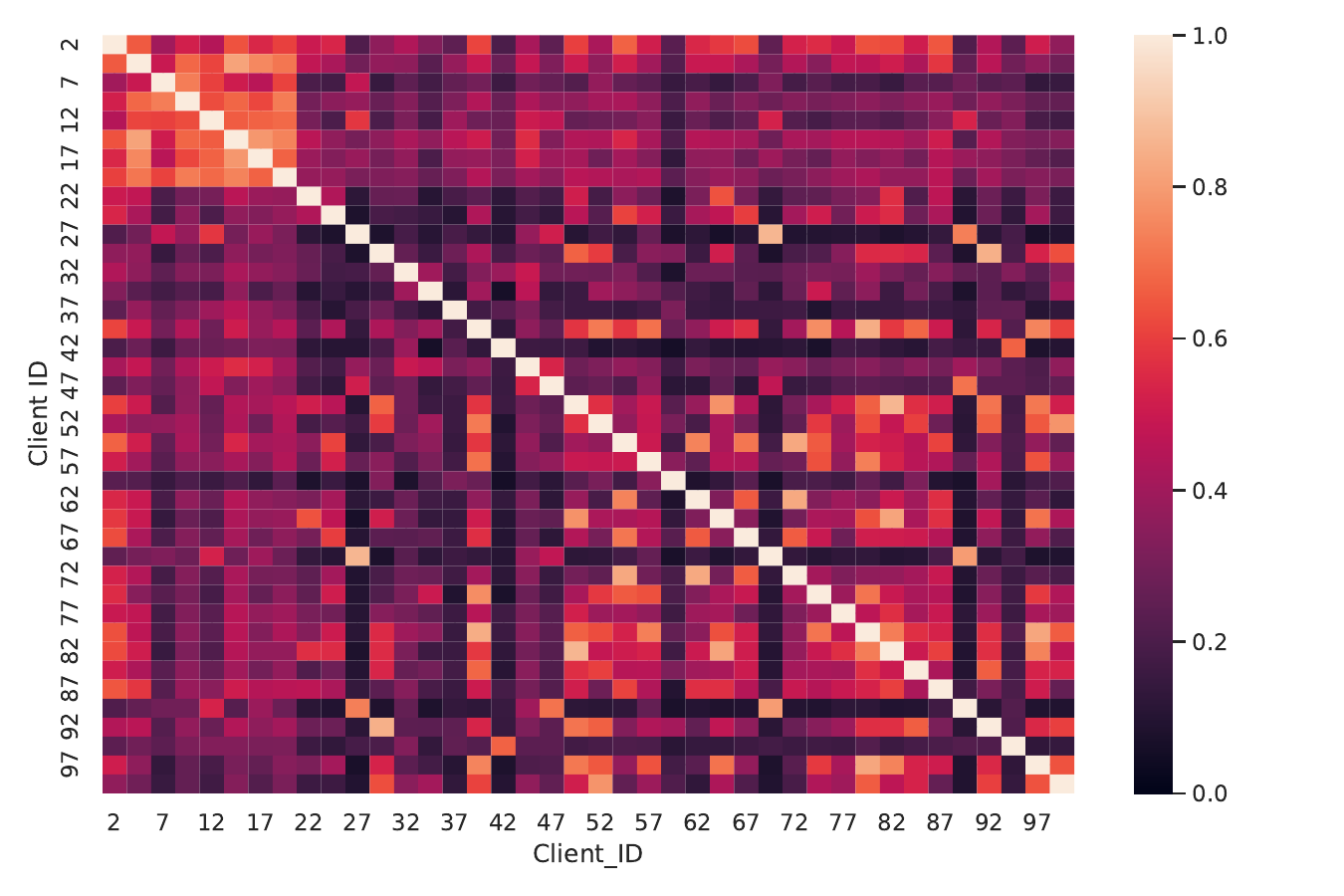}%
\label{fig:cka_inclusivefl_s3}}

\subfloat[Stage 2 of FedIN.]
{\includegraphics[width=4cm]{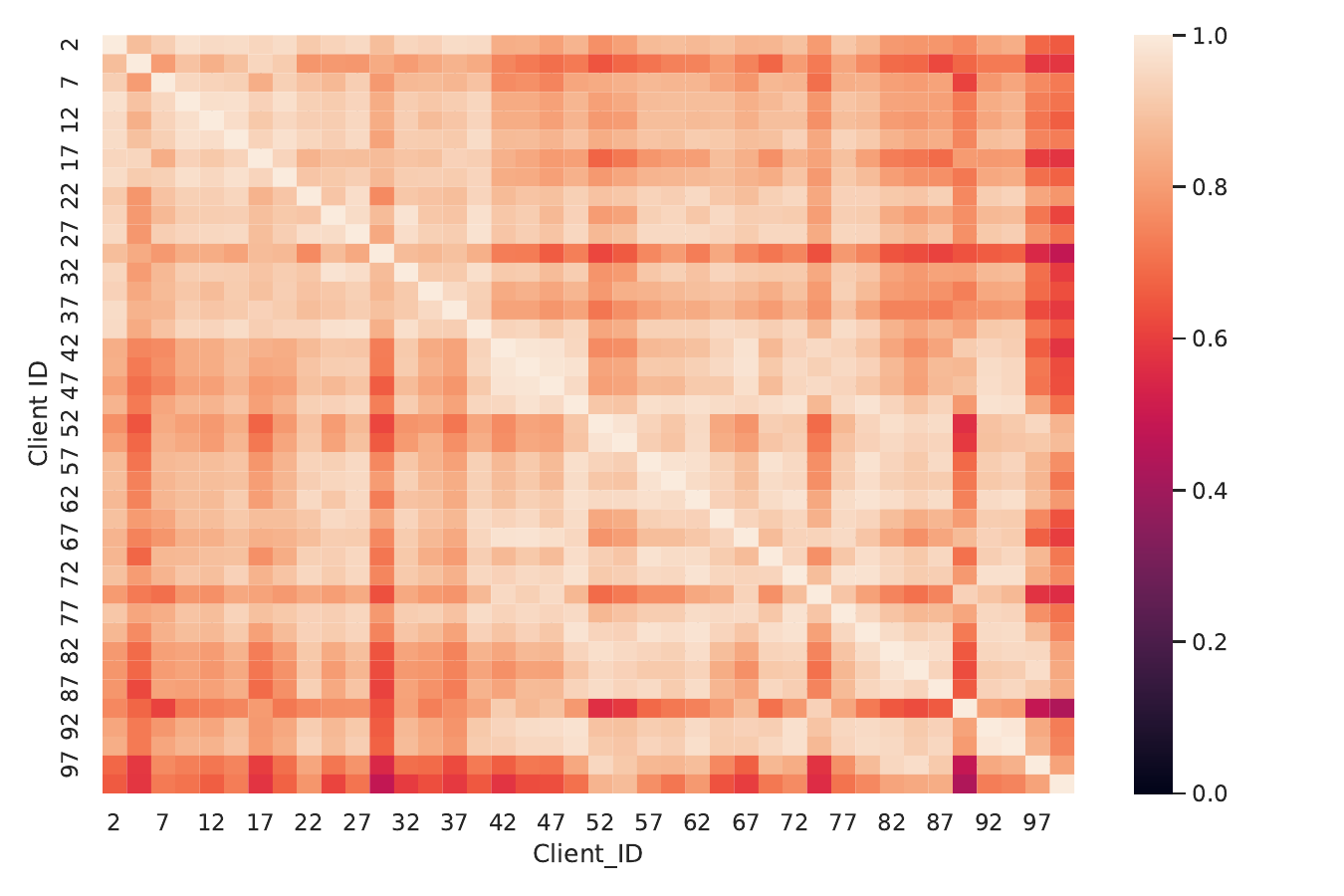}%
\label{fig:cka_fedIN_s2}}
\hfil
\subfloat[Stage 3 of FedIN.]
{\includegraphics[width=4cm]{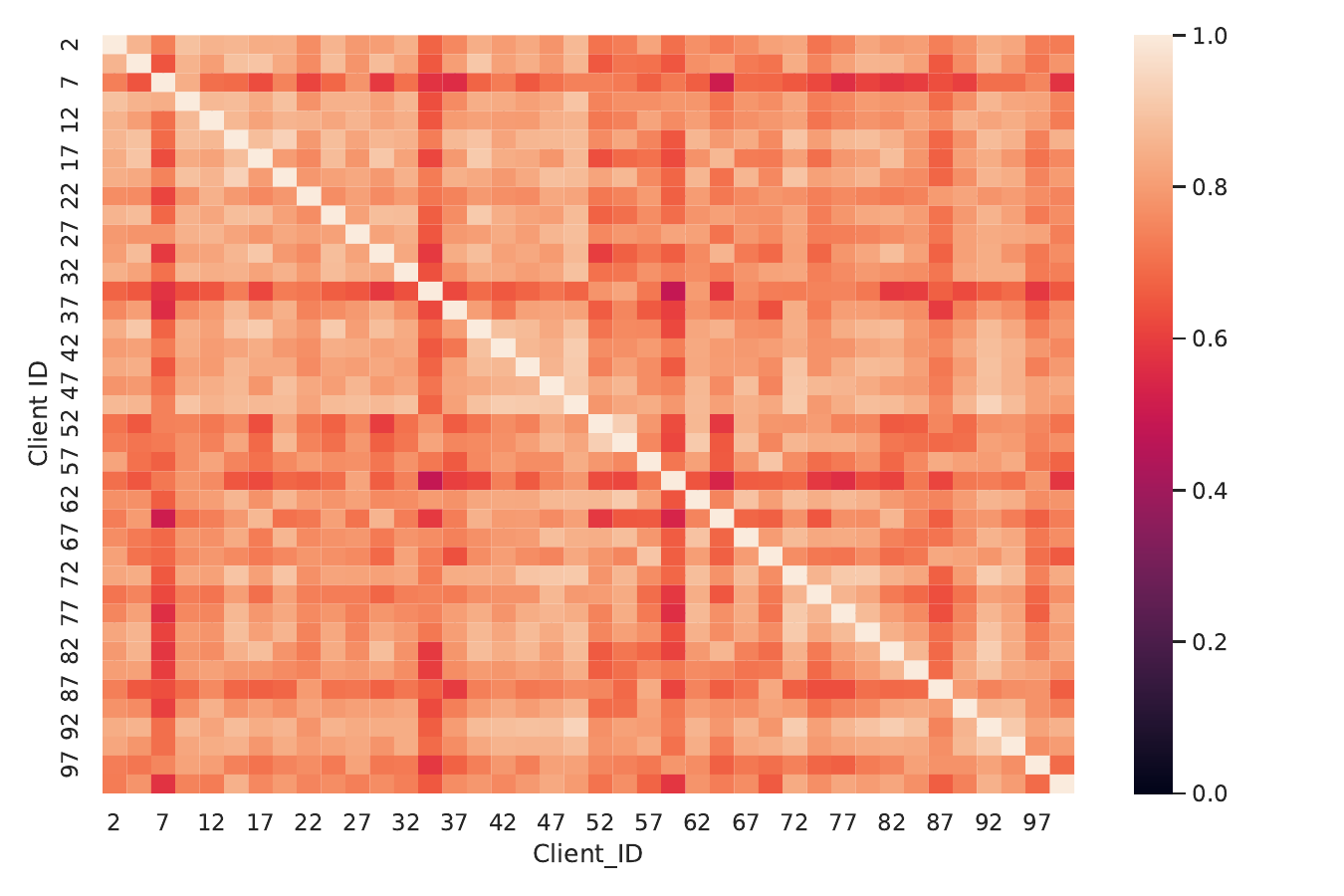}%
\label{fig:cka_fedIN_s3}}

\caption{Heatmaps of CKA similarity from stage 2 and stage 3 among different clients in non-IID data with CIFAR-10.}
\label{fig:cka_heatmaps}
\end{figure}

\begin{figure*}[!t]
\centering
\subfloat[FedAvg.]
{\includegraphics[width=6cm]{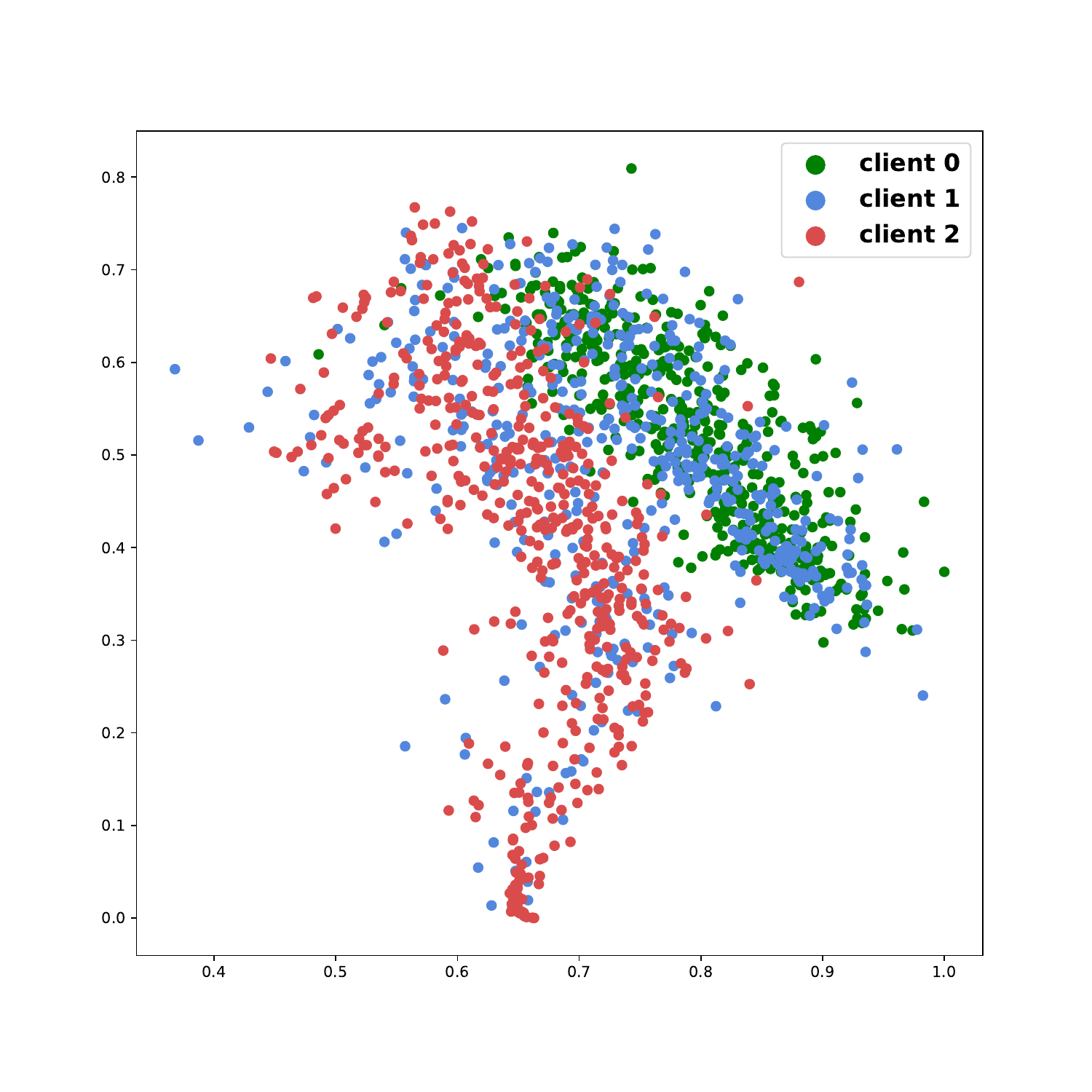}%
\label{fig:tsne_fedavg}}
\hfil
\subfloat[InclusiveFL.]
{\includegraphics[width=6cm]{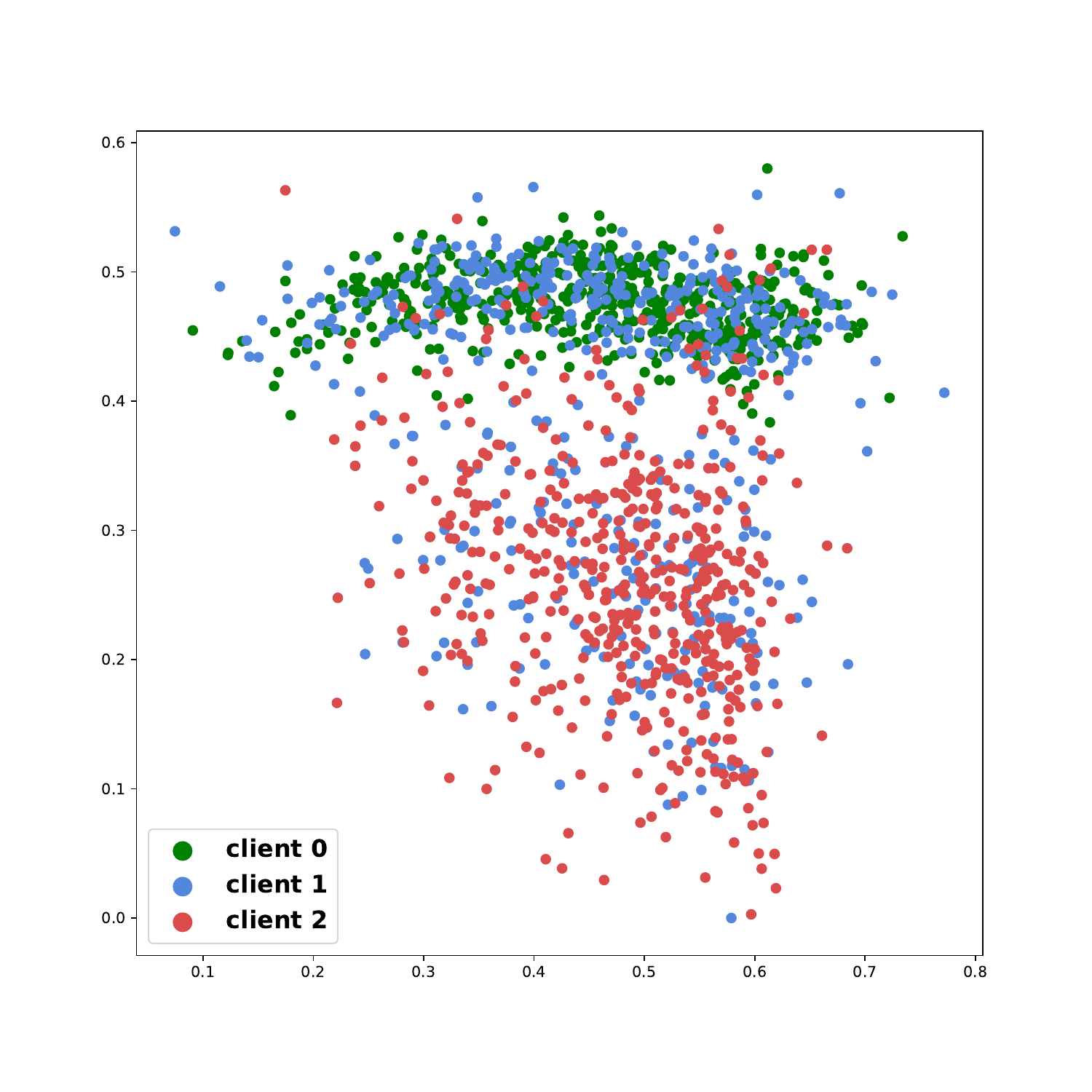}%
\label{fig:tsne_inclusivefl}}
\hfil
\subfloat[FedIN.]
{\includegraphics[width=6cm]{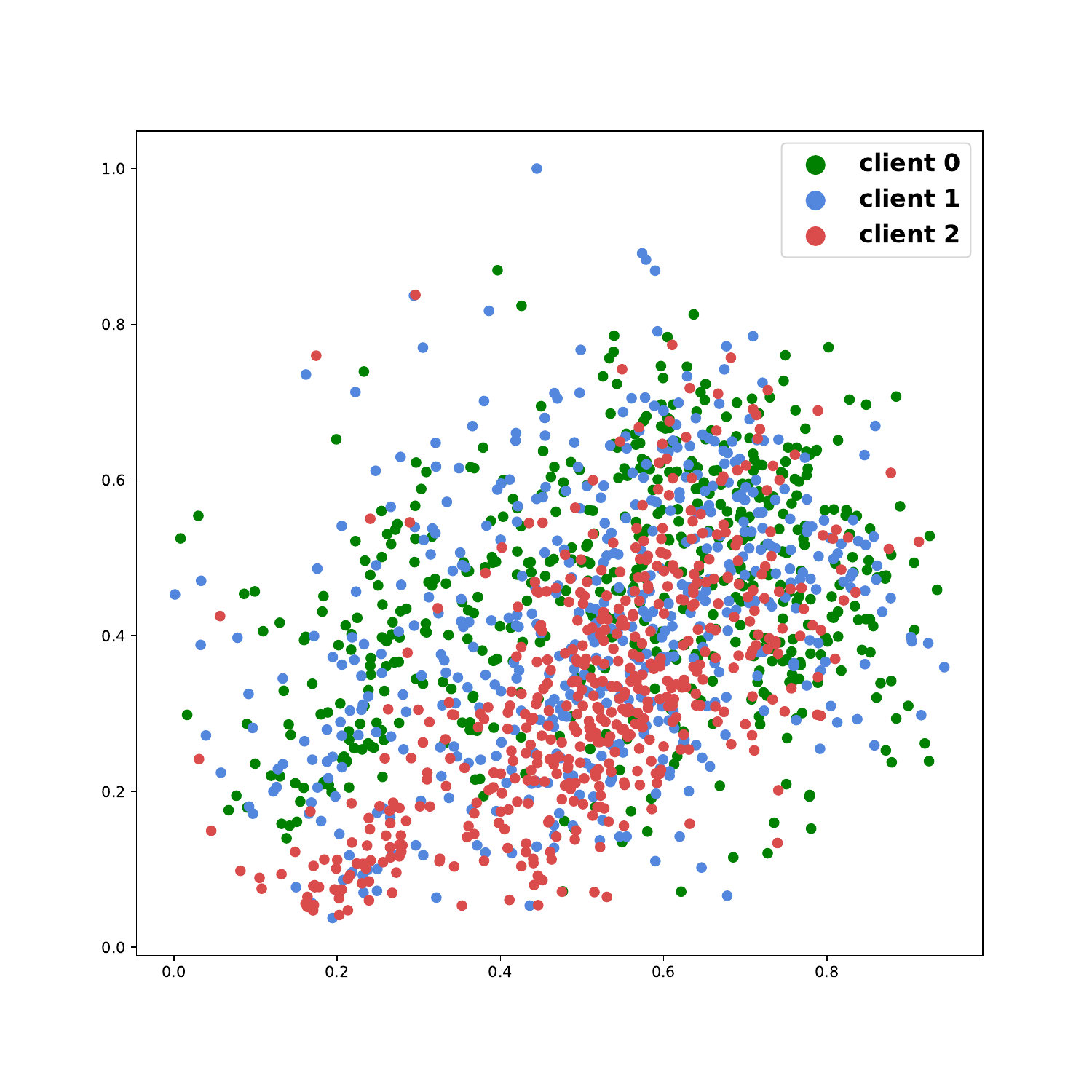}%
\label{fig:tsne_fedIN}}
\caption{t-SNE visualization of features learned by different methods from stage 3 on CIFAR-10. We select data from the same class and utilize three models with different architectures (Client0: ResNet10, Client1: ResNet14, Client2: ResNet26).}
\label{fig:tsne}
\end{figure*}

\subsection{The Reason for the Improvements}

\subsubsection{CKA similarity for different stages}
Inspired by \cite{luo2021no} and \cite{raghu2021vision}, we use CKA similarity \cite{kornblith2019similarity} to examine the layer similarity among different clients across different methods, in order to shed light on the reasons behind the improvements observed with FedIN. A higher CKA similarity indicates that client models can effectively capture common features within the context of system heterogeneity. In our analysis, stage $i$ indicates the $i_{th}$ block in the ResNet architecture, aligned with the corresponding layer $i$ in the PyTorch ResNet source code. It is worth mentioning that ResNet10, ResNet14, ResNet18, ResNet22, and ResNet 26 consist of four stages. Our focus lies on evaluating the CKA similarity of outputs across these four stages. To simplify the figure annotations, we concentrate on three specific methods: FedAvg as an essential baseline, InclusiveFL as a representative method for system heterogeneity, and our proposed method, FedIN.

\figurename~\ref{fig_CKA} illustrates the CKA similarity of different stages under IID and non-IID. Notably, in \figurename~\ref{fig_CKA_iid}, FedIN exhibits the highest similarity even in the deepest stage (stage 3), while FedAvg and InclusiveFL struggle to maintain high similarity levels in stage 3, as evidenced by the gray area in the figure. In \figurename~\ref{fig_CKA_non_iid}, FedIN still maintains a higher similarity than FedAvg and InclusiveFL, especially in the deep stage (stage 3). 

To gain further insights into the dissimilarities between FedIN and the other methods, we present heatmaps of similarity from stage 2 and stage 3 among clients in \figurename~\ref{fig:cka_heatmaps}. \figurename~\ref{fig:cka_fedavg_s2}, \figurename~\ref{fig:cka_inclusivefl_s2}, and \figurename~\ref{fig:cka_fedIN_s2} demonstrate the heatmaps from stage 2. Similar to the observations in \figurename~\ref{fig_CKA_non_iid}, the average similarity of FedIN marginally surpasses that of FedAvg and InclusiveFL. However, in stage 3, as shown in \figurename~\ref{fig:cka_fedavg_s3}, \figurename~\ref{fig:cka_inclusivefl_s3}, and \figurename~\ref{fig:cka_fedIN_s3}, the heatmap corresponding to FedIN (\figurename~\ref{fig:cka_fedIN_s3}) exhibits significantly lighter shades compared to the heatmaps of FedAvg (\figurename~\ref{fig:cka_fedavg_s3}) and InclusiveFL (\figurename~\ref{fig:cka_inclusivefl_s3}). These results and analyses suggest that FedIN ensures consistency among the deep layers of client models, captures valuable shared features, and achieves superior accuracy in the presence of system heterogeneity.

\subsubsection{t-SNE visualization}
We conduct t-SNE visualizations \cite{van2008visualizing} on features extracted from stage 3 in \figurename~\ref{fig:tsne}, focusing on data belonging to the same class. The objective is to observe the clustering behavior of these data points. In \figurename~\ref{fig:tsne_fedavg} and \figurename~\ref{fig:tsne_inclusivefl}, it is evident that the features from client 0 and client 1 and features from client 2 are separated. However, the features from these three clients form a singular cluster in FedIN, as depicted in \figurename~\ref{fig:tsne_fedIN}, validating that the features from data with the same class from different model architectures are consistent in FedIN.

\begin{table}[!t]
    \def\arraystretch{1.6}
    \caption{Model accuracy with ablation studies.}
    \label{tab:acc_CIFAR10_ablation}
    \centering
    \begin{tabular}{cccccc}
    \hline
        \multirow{2}*{CIFAR-10} & \multicolumn{5}{c}{Methods} \\
        \cline{2-6}
        & FedAvg & w/o IN & w/o Prox & w/o Opt & \mycc FedIN  \\
      \hline
       
        IID & 76.8 & 77.6 & 78.8 & 79.4 &\mycc \textbf{80.5} \\

        Non-IID & 66.2 & 72.0 & 66.4 & 74.9 & \mycc \textbf{75.9}\\

    \hline
    \end{tabular}
\end{table}

\begin{figure}[!t]
\centering
\includegraphics[width=8.3cm]{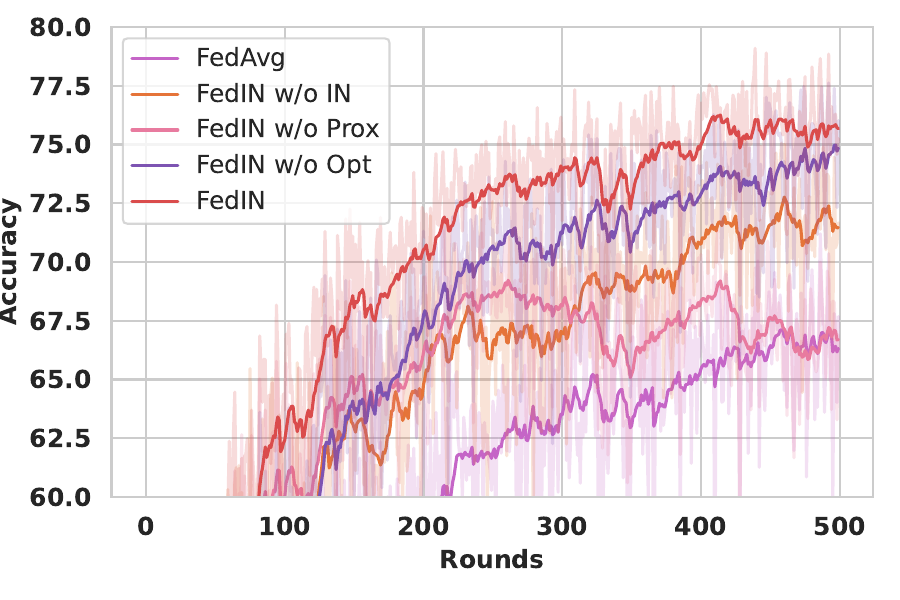}%
\caption{Smoothed test accuracy for non-IID data of CIFAR-10 in the ablation study. The original results of accuracy are the grey lines.}
\label{fig:acc_noniid_cifar10_ablation}
\end{figure}

\begin{figure}[!t]
\centering
\subfloat[Effects from different batch sizes.]
{\includegraphics[width=4cm]{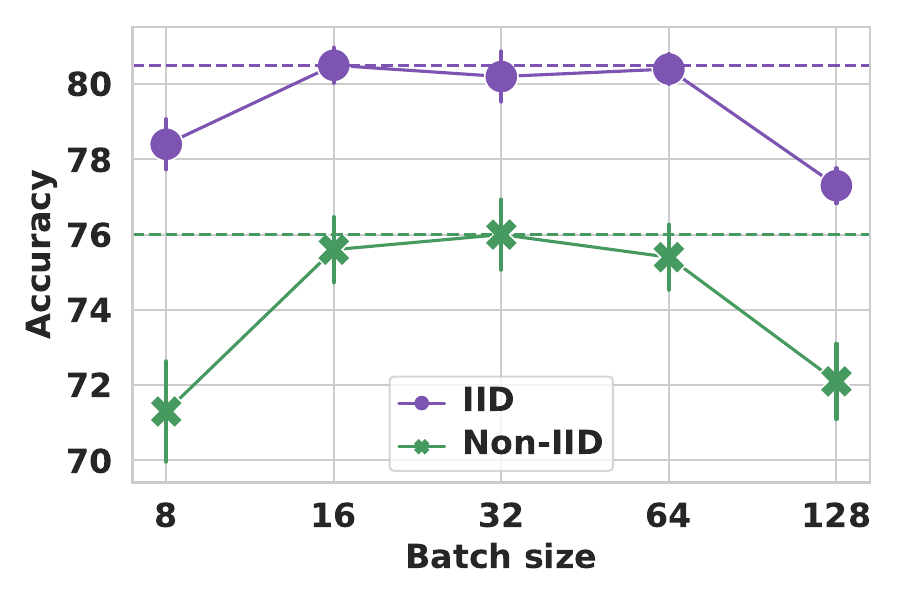}%
\label{fig:batchSize}}
\hfil
\subfloat[Effects from different sample numbers.]
{\includegraphics[width=4cm]{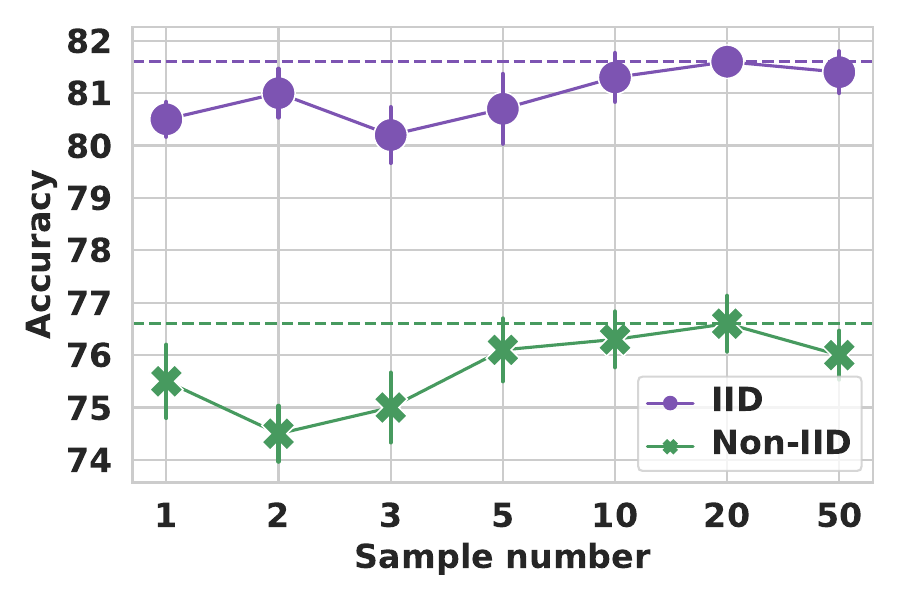}%
\label{fig:sampleNumbers}}
\caption{Illustrations for effects from different batch sizes and different sample numbers in non-IID CIFAR-10.}
\label{fig:robustness}
\end{figure}

\begin{table*}[!t]
    \def\arraystretch{1.6}
    \caption{Model accuracy with different client numbers on CIFAR-10.}
    \label{tab:acc_CIFAR10_different_client_numbers}
    \centering
    \begin{tabular}{ccccccccccc}
    \hline
        \multirow{2}*{Methods} & \multicolumn{5}{c}{IID} & \multicolumn{5}{c}{Non-IID}\\
        \cline{2-11}
        & $N_c=10$ & $N_c=20$ & $N_c=50$ & $N_c=100$ &  $N_c=200$ & $N_c=10$ & $N_c=20$ & $N_c=50$ & $N_c=100$ &  $N_c=200$ \\
      \hline
       
        FedAvg & 79.3 & 79.2 & 78.7 & 76.8 & 74.0 & 68.3 & 67.9 & 66.9 & 66.2 & 62.5\\

        InclusiveFL & 77.5 & 76.7 & 79.1 & 75.0 & 73.4 & 66.8 & 68.4 & 67.1 & 66.1 & 61.2 \\

        \mycc FedIN & \mycc \textbf{82.8} & \mycc \textbf{83.1} & \mycc \textbf{81.0} & \mycc \textbf{80.5} & \mycc \textbf{74.3} & \mycc \textbf{76.7} & \mycc \textbf{76.3} & \mycc \textbf{74.1} & \mycc \textbf{75.9} & \mycc \textbf{72.2} \\

    \hline
    \end{tabular}
\end{table*}

\subsection{Ablation Study}
We conduct an ablation study to evaluate the contributions of the key components in FedIN. Our ablation study includes the following methods: (i) FedAvg, (ii) FedIN w/o IN (FedIN without IN loss), (iii) FedIN w/o Prox (FedIN without Prox regularized term), (iv) FedIN w/o Opt (FedIN without the gradient alleviation (optimization)).
\tablename~\ref{tab:acc_CIFAR10_ablation} and \figurename~\ref{fig:acc_noniid_cifar10_ablation} illustrates the results of the ablation studies. 

\subsubsection{Effects of the gradient alleviation} 
In section \ref{Gradient_divergence}, we propose a convex optimization formulation to address the gradient divergence. In FedIN, we simultaneously update the intermediate layers by the gradients from the local training and IN training by Eq. \ref{updated_gradients}.  This approach, referred to as gradient alleviation, serves as a solution to the gradient divergence problem. In this experiment, we highlight that our solution is advantageous and effective in solving the gradient divergence problem.

\figurename~\ref{fig:acc_noniid_cifar10_ablation} illustrates the results of considering the gradient divergence problem and ignoring this problem. When FedIN disregards this problem, it updates the entire model from the local training, and then continues to update the intermediate layers from the IN training, as depicted by the result of FedIN w/o Opt in \figurename~\ref{fig:acc_noniid_cifar10_ablation}. 
The accuracy achieved by FedIN surpasses that of FedIN without gradient alleviation (FedIN w/o Opt), and the convergence speed of FedIN is also accelerated, as observed in Figure \ref{fig:acc_noniid_cifar10_ablation}. Furthermore, the application of gradient alleviation leads to a performance improvement of 1\% from \tablename~\ref{tab:acc_CIFAR10_ablation}. These findings validate the effectiveness and efficiency of our proposed solution to the gradient divergence problem. It is noteworthy that gradient alleviation does not impose any additional burdens on either the clients or the server.

\subsubsection{Effects of the loss function} In FedIN, the loss function (Eq. \ref{updated_gradients}) incorporates two additional terms, one is IN loss and the other one is Prox regularized term. When considering FedIN w/o Prox, the client models are trained without regularization, and the convergent speed is similar to FedIN w/o IN before 200 rounds as shown in \figurename~\ref{fig:acc_noniid_cifar10_ablation}. However, after 200 rounds, FedIN w/o Prox becomes unstable and its performance deteriorates during the subsequent training process, suggesting that the client models are overfitting to their local dataset. At last, FedIN w/o Prox only achieves the performance like FedAvg, as shown in \tablename~\ref{tab:acc_CIFAR10_ablation}, hinting that the function of IN loss is eliminated at the end of the training process. Therefore, the inclusion of a regularized term becomes essential to maintain the effectiveness of IN loss throughout the training process. After adding the Prox regularized term, FedIN w/o Opt achieves better results than FedIN w/o In and FedIN w/o Prox, indicating the efficiency of the combination of IN loss and Prox regularized term.

\subsubsection{Effects of client numbers}
To investigate the effects of varying client numbers, we conduct experiments on CIFAR-10, as presented in \tablename~\ref{tab:acc_CIFAR10_different_client_numbers}. $N_c$ denotes the number of clients. Notably, FedIN outperforms the other methods across all settings of different client numbers. Furthermore, as the number of clients increases, we observe a decline in accuracy because the amount of local data for each client is also decreased. In the context of a higher number of clients, such as 200 clients, the clients face greater challenges in learning meaningful features due to the limited local data. However, FedIN still achieves 72.2\% with 200 clients under non-IID data, surpassing the performance of FedAvg (62.5\%) and InclusiveFL (61.2\%). These results demonstrate the robustness and effectiveness of the FedIN method in handling the challenges posed by a high number of clients.

\subsubsection{Effects of batch sizes and sample numbers}
We also conduct analysis on different batch sizes and sample numbers on CIFAR-10 to verify the effects of these hyperparameters. As shown in \figurename~\ref{fig:batchSize}, batch sizes 16, 32, and 64 are the best selections, but the batch sizes of 8 and 128 still outperform HeteroFL and InclusiveFL. Considering the communication overhead, a batch size of 16 is the optimal choice. From \figurename~\ref{fig:sampleNumbers}, it is clear that increasing the sample numbers has little impact on accuracy improvement. However, even with a sample number of 1, there is a significant improvement compared to the baselines. It is unnecessary to add excessive overheads to achieve marginal improvement.

\section{Conclusions}
We propose a novel method, called FedIN, which supports model heterogeneity in FL environment. 
FedIN conducts local training based on the private dataset and IN training from the client features, requiring only one batch of features. Moreover, we formulate a convex optimization problem to tackle the gradient divergence problem induced by a combination of local training and IN training. We conduct extensive experiments on IID data and non-IID data from three public datasets with seven baselines. The experiment results illustrate that FedIN achieves superior performances in both IID data and non-IID data. Moreover, we conducted an analysis to elucidate the efficiency and effectiveness of FedIN in heterogeneous environments.
Furthermore, we investigated the contributions of each component of FedIN in the ablation studies. These studies not only highlight the advantages of our proposed solution for addressing the gradient divergence problem but also emphasize the importance of IN training and the impact of varying batch sizes, sample numbers, and client numbers.

\bibliographystyle{IEEEtran}
\bibliography{refer}

\vfill

\end{document}